\documentclass[10pt,twocolumn,letterpaper]{article}

\usepackage{cvpr}      

\definecolor{cvprblue}{rgb}{0.21,0.49,0.74}
\usepackage{flushend}
\usepackage[accsupp]{axessibility}
\usepackage[pagebackref,breaklinks,colorlinks,allcolors=cvprblue]{hyperref}
\usepackage{multirow}
\usepackage{graphicx}
\newcommand{\mycite}[1]{\textcolor{cvprblue}{#1}}

\def\paperID{3770} 
\def\confName{CVPR}
\def\confYear{2025}

\title{Commonsense Video Question Answering through Video-Grounded \\ Entailment Tree Reasoning}

\author{Huabin Liu\textsuperscript{1,3*},\quad Filip Ilievski\textsuperscript{2}, \quad Cees G. M. Snoek\textsuperscript{3} \\
\textsuperscript{1} Shanghai Jiao Tong University~\quad
\textsuperscript{2} Vrije Universiteit Amsterdam~\quad
\textsuperscript{3} University of Amsterdam \\
{\tt\small huabinliu@sjtu.edu.cn,~f.ilievski@vu.nl,~c.g.m.snoek@uva.nl}
}

\begin{document}
\maketitle
\let\thefootnote\relax\footnotetext{$*$\ Work was done as a visiting student at University of Amsterdam}
\begin{abstract}
This paper proposes the first video-grounded entailment tree reasoning method for commonsense video question answering (VQA). Despite the remarkable progress of large visual-language models (VLMs), there are growing concerns that they learn spurious correlations between videos and likely answers, reinforced by their black-box nature and remaining benchmarking biases. Our method explicitly grounds VQA tasks to video fragments in four steps: entailment tree construction, video-language entailment verification, tree reasoning, and dynamic tree expansion. A vital benefit of the method is its generalizability to current video- and image-based VLMs across reasoning types.
To support fair evaluation, we devise a de-biasing procedure based on large-language models that rewrites VQA benchmark answer sets to enforce model reasoning. Systematic experiments on existing and de-biased benchmarks highlight the impact of our method components across benchmarks, VLMs, and reasoning types.
\end{abstract}    
\section{Introduction}
\label{sec:intro}
\vspace{-0.5em}

This paper proposes a video-grounded reasoning method for commonsense video question answering (VQA). VQA has a long tradition in computer vision~\cite{nextqa,intentqa,VIM2023,dibs2024}, with remarkable recent progress obtained through video- and image-language models \cite{videollama,videollava,videochat2,blip2,llava} (throughout this paper collectively referred to as vision-language models, or VLMs). Yet, there are growing concerns that their improved performance is based on learning shortcut associations between videos and likely answers, as opposed to reasoning~\cite{nextgqa}. Such concerns are reinforced by the black-box nature of these models~\cite{videochat2,videollava}, which prohibits a deeper understanding of their decision-making process.

We are inspired by recent work in natural language processing, where entailment trees have emerged as a mechanism to explicitly analyze answer candidates, using LLMs to recursively decompose a candidate into hypotheses and natural language inference formalisms to evaluate the hypotheses \cite{entailmentbank}.
Entailment trees provide an explicit reasoning chain that explains the model's decision-making process and enables verification of each step, thus addressing concerns about shortcut learning. Recently, Sanders \etal~\cite{tvtree} have devised a mechanism to apply entailment trees to videos. However, their work assumes video transcripts 
are explicitly provided to evaluate answers, thus avoiding the complexity of grounding hypotheses into video content. In this paper, we propose the first video-grounded entailment tree reasoning method for commonsense VQA.

\begin{figure}
    \centering
    \includegraphics[width=\linewidth]{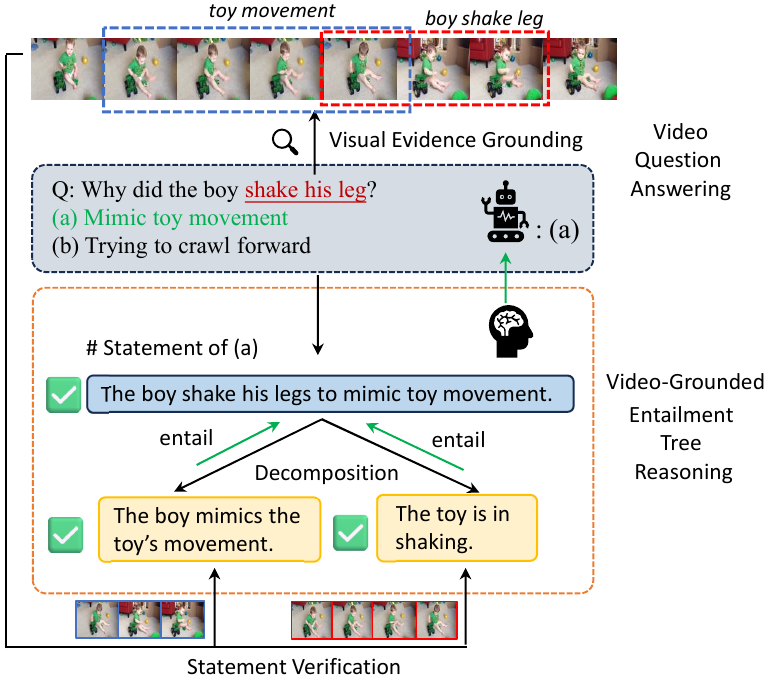}
    \caption{Given a video questioning answering task, our framework performs explicit reasoning over an entailment tree, where answer options are transformed into statements. These statements are then recursively decomposed and verified based on video-grounded evidence relevant to the question.}
    \label{fig:intro}
        \vspace{-0.5em}
\end{figure}

Our method explicitly grounds VQA tasks to video fragments in four steps: (\textit{i}) entailment tree construction, (\textit{ii}) video-language entailment verification, (\textit{iii}) tree reasoning, and (\textit{iv}) dynamic tree expansion. As shown in~\cref{fig:intro}, given a video and a multiple-choice question, we generate a statement for each answer candidate that acts as a first-level hypothesis. We decompose each statement iteratively, aiming to produce sub-statements that can be confidently verified in the video. The video is itself decomposed into partitions, consisting of sets of frames. Verifying each statement is then a matter of aligning it to a video partition. 
A vital benefit of the method is its generalizability to current video and image-based VLMs across reasoning types, including temporal and causal. To demonstrate its video reasoning ability, we develop an answer-set de-bias procedure supported by an LLM that ensures that VQA benchmarks~\cite{nextqa,intentqa} are adequate for reasoning in videos without relying on spurious correlations.
Our experiments show that our video-grounded entailment tree method consistently improves video- and image-based baselines on both the existing and de-biased benchmarks. Moreover, it performs on par with, and often better than, state-of-the-art video-based VLMs while leveraging $257\times$ fewer parameters. Further ablations show that the method benefits from considering both textual and video information and that its performance is especially strong on causal and temporal questions.

\section{Related work}
\vspace{-0.5em}

\noindent\textbf{Video question answering.}

Recent research has shown that while video-based VLMs can achieve state-of-the-art performance, their answers are sensitive to object size, positioning, and speed~\cite{yang2023mitigating,tot}. Moreover, when answering temporal and spatial questions, VLMs rely on textual biases to ``guess'' answers rather than performing genuine understanding and reasoning over visual-text information~\cite{zhang2024can}.

To improve the robustness and interpretability of VLMs, one line of research enriches VLMs with visual grounding functionality during QA, which enables VLMs to localize relevant video moments~\cite{timecraft,nextgqa,seviLA} or key frames~\cite{videotree,task2022} to support answers. 
However, while these methods localize visual evidence, the process by which VLMs use it to deduce answers remains opaque. Another approach leverages external LLMs as reasoners or agents to enhance interpretability in textual modality. For instance, LLoVi~\cite{llovi} converts VQA to a text-based QA task via video captioning, then prompts an LLM to provide answers. Similarly, VideoAgent~\cite{videoagent} uses an LLM to recursively determine if the current frames can answer the given question based on their textual descriptions. However, these methods heavily rely on the reasoning capabilities of LLMs. Like VLMs, the LLM reasoning process remains a black box, and hallucinations are common. Recently, TV-trees~\cite{tvtree} attempted to perform explicit reasoning over both visual and textual modalities using a neuro-symbolic system. However, their work assumes video transcripts 
are explicitly provided to evaluate answers, thus avoiding the complexity of grounding hypotheses into video content. Instead, we contribute a general framework for explicit reasoning in commonsense VQA, fueled by a grounding component that aligns question components with video fragments.

Beyond methodology, some works focus on providing fair and comprehensive VLM evaluations in VQA tasks by creating new benchmarks~\cite{videomme,benchmark1,mecd2024}. These benchmarks contain videos with diverse scenarios and durations, with carefully crafted questions and options designed to prevent textual shortcuts that VLMs might exploit. Video-specific questions (e.g., compositional action reasoning)~\cite{tot}, which require insights beyond textual associations, are included to test commonsense reasoning in VLMs. Addressing concerns of remaining biases in such benchmarks~\cite{videomme}, we contribute an LLM-based answer-set de-biasing procedure to ensure that VQA benchmarks~\cite{nextqa,intentqa} are adequate to evaluate reasoning in videos rather than spurious correlations.

\noindent\textbf{Systematic language reasoning.}
As LLMs demonstrate great potential in reasoning, there has been considerable interest in using LLMs to generate systematic explanations to support their answers. The series of Chain-of-Thought  prompting~\cite{cot,tree-of-thought,graph-of-thought} encourages LLMs to think step-by-step to perform explicit multi-hop reasoning, providing free-form reasoning steps before arriving at an answer. However, such implicit explanations are not grounded in external knowledge or evidence, which may lead to unverifiable and unfaithful reasoning. Since the development of EntailmentBank~\cite{entailmentbank}, research has increasingly focused on constructing explanation trees~\cite{entailer,nellie} and graphs~\cite{rationality}, encouraging models to generate step-wise entailment proofs of a statement using a set of supporting facts. Entailer~\cite{entailer} introduced this systematic explanation framework into language-based multiple-choice QA, performing explicit reasoning by generating entailment trees grounded in the model’s internal beliefs. REFLEX~\cite{rationality} extends the entailment tree to form a belief graph for QA models, aiming to address consistency issues by intervening in the intermediate reasoning steps. Instead of grounding facts in predefined rules or model beliefs, NELLIE~\cite{nellie} adopts Prolog-based inference engines and external natural language corpora to build entailment trees as explainable reasoning for multiple-choice QA tasks. While such techniques for natural language processing inspire our framework, we generalize entailment trees to VQA, contributing a novel grounding method that aligns entailment trees with video fragments.

\section{Video-grounded entailment tree reasoning}
\label{sec:method}
\begin{figure*}
    \centering
    \includegraphics[width=1.0\linewidth]{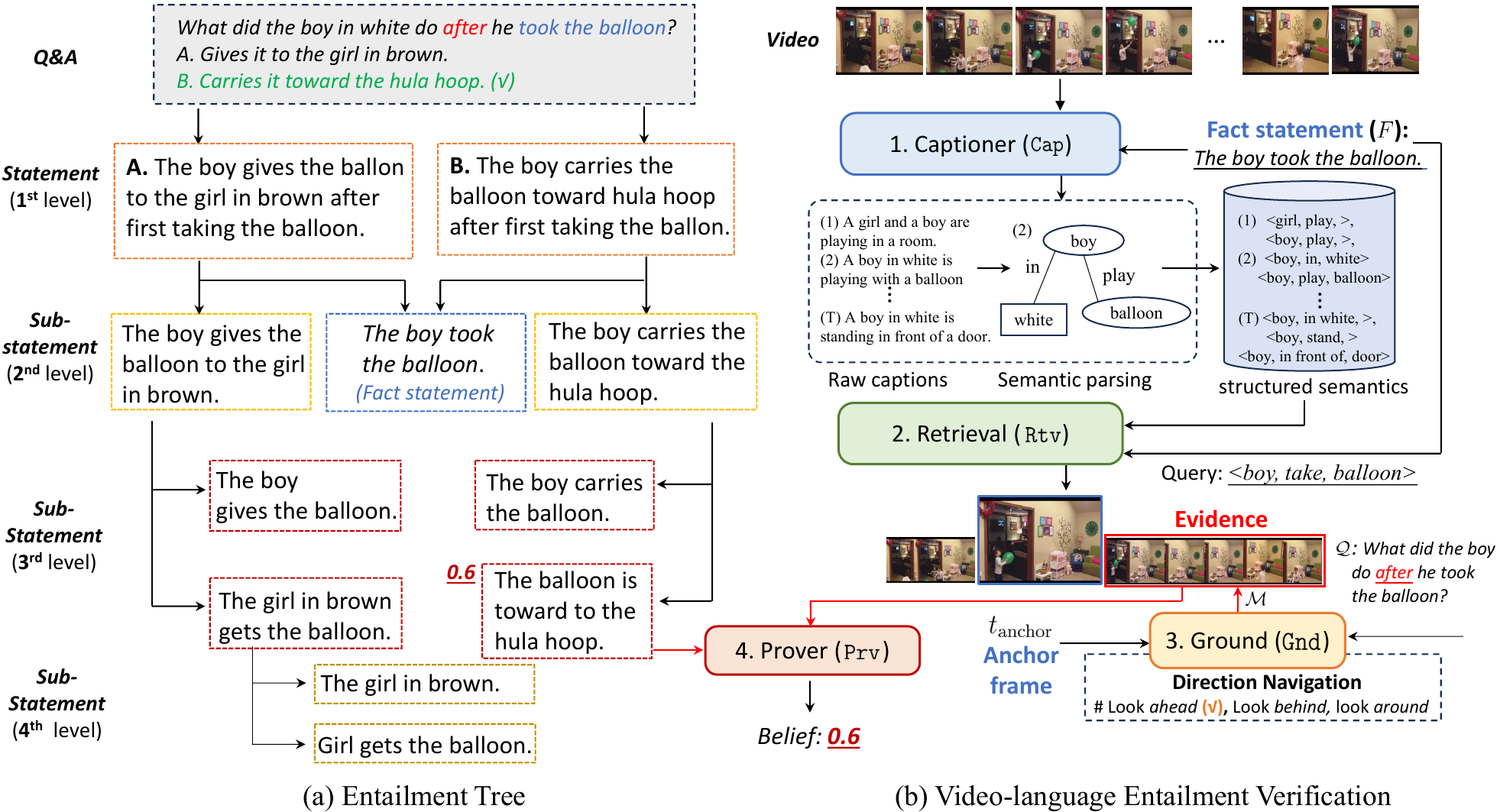}
    \caption{Overview of our framework. (a) The generation of the entailment tree, where statements are recursively decomposed until the tree reaches its max depth or meets the stop criterion. (b) The process of video-language entailment verification: the input video is first converted into textual descriptions. Each caption is then parsed into structured semantics. Given the fact statement as a query, we retrieve the anchor frame. Then, based on the temporal or causal navigation indicated by questions, the visual evidence moment can be grounded.}
    \label{fig:overall}
    \vspace{-0.5em}
\end{figure*}

\vspace{-0.5em}
This paper devises a novel explainable framework for grounded commonsense VQA. It derives the answers through systematic reasoning over video-text information with entailment trees. Specifically, in the entailment tree (\cref{fig:overall}a), each candidate answer is decomposed into statements that entail the answer, explaining why each answer could be plausible. These statements are then grounded in relevant visual evidence from the video to prove or refute them (\cref{fig:overall}b). While entailment trees in natural language processing are constructed based on a model's internal knowledge or corpora~\cite{entailer,rationality,nellie}, we ground entailment trees into video fragments. 
Finally, backtracking through the entailment tree leads to systematic reasoning over the statements (\cref{fig:dynamic}). Thus, answers can be deduced by a systematic structure with explicit reasoning paths and explanations rather than relying on opaque, black-box models. 

\subsection{Entailment tree construction}
\label{sec:tree}
\vspace{-0.5em}

\noindent \textbf{Initial statement generation.} Given a question and its answer candidates, we first convert each question-answer pair into a \textit{declarative} sentence that preserves the semantic meaning of the original QA pair. As a result, an $N$-way multiple-choice QA problem produces a set of statements, denoted as $\mathcal{D} {=}{d_1, \dots, d_N}$. For example, the two-way question ``\textit{What did the boy in white do after he first took the balloon? (A) resting on a chair (B) carries it toward the hula hoop}'' is transformed into:
$\mathcal{D}:\{d_1 {=} $ ``\textit{The boy in white resting on a chair after first taking the balloon.}", $d_2 {=} $ ``\textit{The boy in white carries it toward the hula hoop after first taking the balloon.}'' $\}$.
Thus, selecting the best answer equals identifying the correct statement for a given video. 

\noindent \textbf{Recursive statement decomposition.} For each initial statement in $\mathcal{D}$, we generate two sub-statements as proofs that support the statement: 
\texttt{Statement} $\Leftarrow$ \texttt{Sub-statement1, Sub-statement2}. 
The statement is \texttt{True} if and only if both its sub-statements are proved to be \texttt{True}, i.e., the sub-statements \textit{entail} the statement. Proving the original statements is thus translated into proving two simpler sub-statements. This procedure is recursive: the sub-statements can be further decomposed into further sub-statements that entail them. Therefore, to construct an entailment tree, we recursively decompose these sub-statements as new statements in the next tree layer until reaching the maximum depth or meeting the stop criterion. \cref{fig:overall}(a) presents an example of entailment tree generation.

We leverage LLM prompting for both the initial statement generation and the statement decomposition, as these are linguistic tasks (see implementation details).

\subsection{Video-language entailment verification}
\label{sec:verification}
\vspace{-0.5em}
Given the entailment tree, the framework then verifies language statements based on the grounded video content as evidence. Specifically, each statement in the entailment tree must be proven or refuted by analyzing the video. A straightforward solution is to encode the whole video to collect information that can be used to verify the statement. However, the critical visual evidence that accurately verifies a statement tends to exist in a local moment instead of the whole video. Therefore, we develop a novel video grounding that guides the verification process to the moments with relevant visual evidence. 

\noindent \textbf{Question-aware video captioning.}  
Given a video, we convert its visual information into detailed textual information. Specifically, we input video frames into a VLM-based \emph{captioner} \texttt{Cap}($\cdot$) to obtain a caption $c_i {=} \texttt{Cap}(f_i)$ for each frame. However, captioning frames individually can overlook essential details or introduce irrelevant information for VQA. 
In commonsense VQA, questions often focus on specific \textit{facts} already observed in the video. For example, a typical temporal reasoning question is “\textit{What happened before/after Event-A?}” where \textit{Event-A} refers to a fact statement about an event in the video. The fact referenced by the question can be leveraged to guide video understanding. To this end, we first extract the anchor fact indicated by the question and provide it to \texttt{Cap}($\cdot$) as prior knowledge, encouraging the generation of relevant captions. Moreover, captions from all previous frames are also provided for each current frame to ensure \texttt{Cap}($\cdot$) captures the temporal context from the past. This process is formulated as:
\begin{equation}
	c_i = \texttt{Cap}(f_i~|~F, (c_1,\cdots,c_{i-1})),
\end{equation}
where \textit{F} indicates the fact statement.

\noindent \textbf{Video evidence grounding.}
For commonsense VQA, depending on how the question reasons around the fact statement, the necessary evidence for answers can be gathered from specific video moments. For instance, in the case of temporal reasoning (e.g., before or after questions), the answer should be inferred from moments occurring either before or after the time of the relevant fact. Following this intuition, we design a two-step evidence-grounding strategy to localize the critical moments for answering. 

\textbf{\textit{First}}, given the frame-wise captions, we retrieve a keyframe deemed most relevant to the fact statement, which we refer to as the \textit{anchor frame}. A straightforward retrieval approach would involve comparing each $c_i$ with the fact description using specific metrics to identify the anchor frame. However, we enhance retrieval accuracy by adopting a structured semantic retrieval strategy. Specifically, the textual descriptions of each frame and fact statement are converted into structured \textit{triplets}. These triplets capture the attributes and relationships of objects in each frame through structured semantics. As shown in~\cref{fig:overall}(a), rather than directly comparing raw textual descriptions of frames and fact statements, we use these triplets for retrieval. Inspired by the success of using LLMs for retrieval tasks, we prompt an LLM to conduct anchor frame retrieval using the triplets of the fact statement as the query. The LLM then identifies and returns the most relevant frame ID, i.e., its timestamp.
\begin{equation}
t_{\text{anchor}} = \texttt{Rtv}(c_i, F),
\end{equation}
where $t_{\text{anchor}}$ is the time stamp of the anchor frame, $\texttt{Rtv}(\cdot)$ denotes the retrieval process. \textbf{\textit{Second}}, we determine the final moment where we should look centered on the $t_\text{anchor}$, to incorporate the temporal relations present in the question. 
Therefore, based on the anchor frame, the navigation for the moment is selected from ``\textit{look ahead, look behind, look around}'' by considering the question:
\begin{equation}
    \mathcal{M} = \texttt{Gnd}(t_{\text{anchor}} | \mathcal{Q}),
\end{equation}
where \texttt{Gnd}($\cdot$) is the grounding process and $\mathcal{M}$ denotes the grounded continuous interval in the video. Then, frames are resampled from the video within $M$ and used as visual evidence proving or refuting entailment tree statements. 

\noindent \textbf{Visual-text statement prover.} 
Given grounded visual evidence $\mathcal{M}$ of the video, statements are estimated to be true or false.
Specifically, we employ a VLM as the statement \emph{prover}, denoted as \texttt{Prv}($\cdot$), to evaluate each statement within the tree by probing VLM's internal belief on this statement. Each statement will be transformed into a binary QA task, with possible options being \texttt{True} or \texttt{False}. We then directly probe the \texttt{Prv}($\cdot$) with the binary QA prompt and use the next token prediction probabilities of the words to elicit the model's belief. We normalize the prediction logits of the two options to get the confidence score of that statement. The above process is formulated as:
\begin{gather}
    s_d = \texttt{Prv}(\mathcal{M}, h), ~s\in[0,1],
\end{gather}
where $\mathcal{M}$ is the grounded moment and $h$ denotes the statement that needs to be verified.

\begin{figure}
    \centering
    \includegraphics[width=\linewidth]{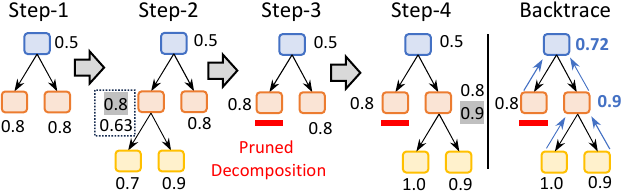}
    \caption{Illustration of dynamic tree generation and backtrace. In Step-3, when the proof score of the left statement calculated from its child nodes is less than its direct score ($0.63<0.8$), its decomposition is pruned and stops.}
    \label{fig:dynamic}
    \vspace{-1em}
\end{figure}

\subsection{Dynamic entailment tree expansion}
\label{sec:dynamic}
\vspace{-0.5em}
So far, we have performed statement decomposition recursively to construct an entailment tree with pre-defined depth. However, not all statements need to be verified recursively, especially those easily determined to be true or false by VLMs. Moreover, as the depth increases, some statement sentences are atomic and directly verifiable. 
Thus, to improve the efficiency of the reasoning process, we further adopt a strategy to expand the entailment tree dynamically. 
Specifically, each statement $d$ is tied with two confidence scores provided by the \texttt{Prv}($\cdot$): 
\begin{itemize}
    \item[(1)] The direct score $s_d$, which indicates the belief of \texttt{Prv}($\cdot$) model in $d$.
    \item[(2)] The proof score $s_p$, denoting how confidently the model can prove $d$, is calculated by multiplying the scores of its direct sub-statements.
\end{itemize}
For a statement $d$, the goal of decomposition is to establish a more reliable and convincing proof path than merely evaluating whether $d$ is true by VLMs. If the decomposition-based reasoning can prove $d$ with higher confidence than its direct score, the overall confidence for statement $d$ should increase. Otherwise, the decomposition should be disregarded. Thus, in the dynamic tree expansion, if decomposition does not enhance a statement’s score, it is pruned, and that statement node becomes a leaf in the entailment tree. \cref{fig:dynamic} presents a toy example. This criterion ensures that only beneficial decompositions are retained, significantly enhancing the tree reasoning process's efficiency. 

\subsection{Reasoning over the entailment tree}
\vspace{-0.5em}
Finally, we perform a backtrace through the entailment tree to calculate the confidence score of each top statement. Specifically, the final score for each statement is produced by comparing its direct score $s_d$ and proof score $s_p$, i.e., $s{=}max(s_d, s_p)$ during backtrace (as shown in~\cref{fig:dynamic}). The overall framework selects the answer corresponding to the statement at the top layer with top-scoring proof.

\section{De-biasing commonsense VQA answer sets}
\vspace{-0.5em}
\begin{figure}
    \centering
    \includegraphics[width=1.0\linewidth]{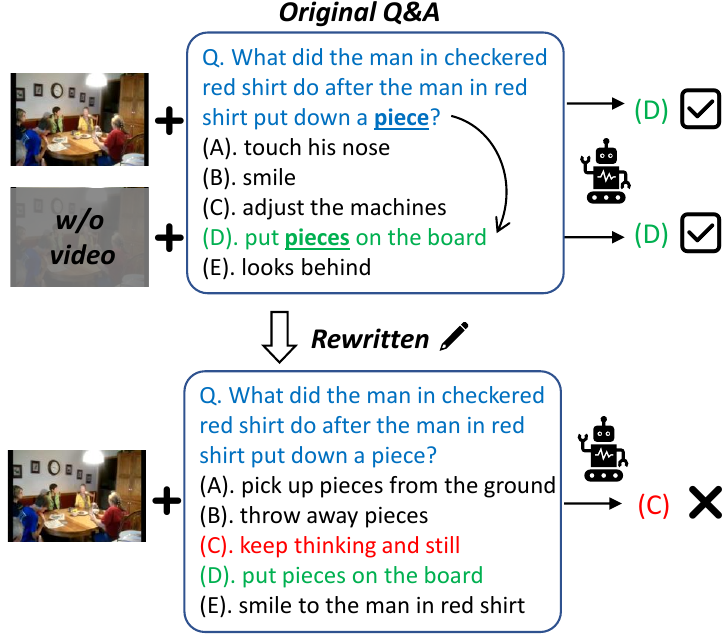}
    \caption{Illustration of commonsense bias in video question answering. The example is selected from the NExT-QA dataset.} 
    \label{fig:commonsense}
    \vspace{-0.8em}
\end{figure}

To demonstrate the reasoning ability of video-grounded entailment trees, it is essential to evaluate using commonsense VQA benchmarks that enforce model reasoning. 
Recent work~\cite{nextgqa,timecraft,invariant} has provided evidence that shortcuts are present in VQA datasets which enables VLMs to solve these tasks based on textual associations rather than video-grounded reasoning. 
While VQA benchmarks increasingly focus on commonsense reasoning skills, such as temporal (e.g. \textit{after, before}) or causal (\textit{how, why, what if}) relationships in video content, reasoning shortcuts affect the validity of their evaluation.
This is illustrated in~\cref{fig:commonsense} (top), where the correct answer (D) is much more relevant to the question and also aligns best with real-world expectations. Consequently, a VLM (VideoLLaVA~\cite{videollava} used in this example) can answer this question correctly by leveraging such associations and without analyzing the video content. Meanwhile, replacing the answer set distractors with other commonsensical answer candidates, as illustrated in~\cref{fig:commonsense} (bottom), makes this task challenging for VLMs. Here, a VLM switches its answer incorrectly to option (C), which confirms the impact of commonsense associations and the lack of grounded reasoning by these models.

\begin{figure}
    \centering
    \includegraphics[width=1.0\linewidth]{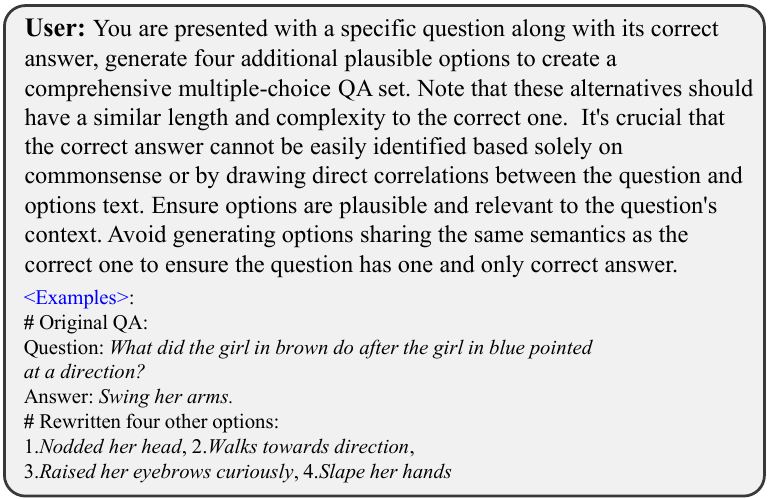}
    \caption{Prompt used for rewriting answers on NExT-QA.}
    \label{fig:rewrite_prompt}
    \vspace{-0.8em}
\end{figure}

To this end, we devise a de-biasing procedure that mitigates reasoning shortcuts in commonsense VQA answer sets. 
Our de-biasing procedure transforms multiple-choice VQA benchmarks (e.g., NExT-QA) by rewriting their answer distractors while keeping their question and ground-truth answer intact. We prompt an LLM (LLaMA-3) to implement the rewriting procedure for each original QA set. \cref{fig:rewrite_prompt} shows the detailed prompt we used for LLaMA-3 on NExT-QA dataset. 
This procedure ensures that (1) the answers cannot be easily derived from the QA set associations and (2) the answer remains consistent with the original QA pair. Thus, our procedure enables the scalable construction of de-biased QA sets by leveraging the commonsense associations in LLMs. 
The next section, focusing on experimental evaluation, analyzes the application of de-biasing to various datasets and its impact on the performance of VLMs, with and without entailment tree reasoning.

\section{Experiments}
\vspace{-0.5em}
\subsection{Experimental setup}
\vspace{-0.5em}
\begin{table*}[t]
\centering
\resizebox{\textwidth}{!}{%
\begin{tabular}{cc|cc|cc|llll|l}
\hline
\multicolumn{2}{c|}{} &
  \multicolumn{2}{c|}{NExT-QA} &
  \multicolumn{2}{c|}{IntentQA} &
  \multicolumn{4}{c|}{VideoMME} &
   \\ \cline{3-10}
\multicolumn{2}{c|}{\multirow{-2}{*}{Model}} &
  Temporal &
  Causal &
  Temporal &
  Causal &
  Temporal &
  Spatial &
  Action &
  Object &
  \multirow{-2}{*}{Avg} \\ \hline
\multicolumn{1}{c|}{} &
  BLIP-2~\cite{blip2} &
  38.3 &
  36.1 &
  43.8 &
  48.6 &
  25.4 &
  26.9 &
  24.2 &
  28.6 &
  34.0 \\
\multicolumn{1}{c|}{} &
  \cellcolor[HTML]{E2E1E1}+Ours &
  \cellcolor[HTML]{E2E1E1}45.3 &
  \cellcolor[HTML]{E2E1E1}41.8 &
  \cellcolor[HTML]{E2E1E1}48.9 &
  \cellcolor[HTML]{E2E1E1}52.5 &
  \cellcolor[HTML]{E2E1E1}30.5 &
  \cellcolor[HTML]{E2E1E1}27.4 &
  \cellcolor[HTML]{E2E1E1}27.7 &
  \cellcolor[HTML]{E2E1E1}28.8 &
  \cellcolor[HTML]{E2E1E1}37.9 \\ \cline{2-11} 
\multicolumn{1}{c|}{} &
  LLaVA-1.5~\cite{llava} &
  37.8 &
  40.7 &
  45.8 &
  50.0 &
  31.1 &
  33.6 &
  27.7 &
  29.8 &
  37.1 \\
\multicolumn{1}{c|}{\multirow{-4}{*}{\begin{tabular}[c]{@{}c@{}}Image-based\\ VLMs\end{tabular}}} &
  \cellcolor[HTML]{E2E1E1}+Ours &
  \cellcolor[HTML]{E2E1E1}45.6 &
  \cellcolor[HTML]{E2E1E1}47.9 &
  \cellcolor[HTML]{E2E1E1}48.4 &
  \cellcolor[HTML]{E2E1E1}54.7 &
  \cellcolor[HTML]{E2E1E1}36.7 &
  \cellcolor[HTML]{E2E1E1}36.8 &
  \cellcolor[HTML]{E2E1E1}31.4 &
  \cellcolor[HTML]{E2E1E1}30.7 &
  \cellcolor[HTML]{E2E1E1}41.5 \\ \hline
\multicolumn{1}{c|}{} &
  VideoChat2~\cite{videochat2} &
  56.9 &
  62.1 &
  60.4 &
  63.2 &
  50.3 &
  52.4 &
  49.5 &
  50.1 &
  55.6 \\
\multicolumn{1}{c|}{} &
  \cellcolor[HTML]{E2E1E1}+Ours &
  \cellcolor[HTML]{E2E1E1}57.8 &
  \cellcolor[HTML]{E2E1E1}61.6 &
  \cellcolor[HTML]{E2E1E1}62.3 &
  \cellcolor[HTML]{E2E1E1}63.8 &
  \cellcolor[HTML]{E2E1E1}52.8 &
  \cellcolor[HTML]{E2E1E1}51.5 &
  \cellcolor[HTML]{E2E1E1}51.3 &
  \cellcolor[HTML]{E2E1E1}50.0 &
  \cellcolor[HTML]{E2E1E1}56.4 \\ \cline{2-11} 
\multicolumn{1}{c|}{} &
  VideoLLaVA~\cite{videollava} &
  56.0 &
  60.4 &
  53.9 &
  60.7 &
  47.6 &
  44.3 &
  46.2 &
  49.7 &
  52.3 \\
\multicolumn{1}{c|}{} &
  \cellcolor[HTML]{E2E1E1}+Ours &
  \cellcolor[HTML]{E2E1E1}58.3 &
  \cellcolor[HTML]{E2E1E1}62.7 &
  \cellcolor[HTML]{E2E1E1}57.6 &
  \cellcolor[HTML]{E2E1E1}61.8 &
  \cellcolor[HTML]{E2E1E1}48.3 &
  \cellcolor[HTML]{E2E1E1}46.1 &
  \cellcolor[HTML]{E2E1E1}49.8 &
  \cellcolor[HTML]{E2E1E1}50.3 &
  \cellcolor[HTML]{E2E1E1}54.2 \\ \cline{2-11} 
\multicolumn{1}{c|}{} &
  VideoLLaMA~\cite{videollama} &
  55.4 &
  60.2 &
  55.1 &
  56.7 &
  44.8 &
  47.2 &
  44.7 &
  49.3 &
  51.7 \\
\multicolumn{1}{c|}{\multirow{-6}{*}{\begin{tabular}[c]{@{}c@{}}Video-based\\ VLMs\end{tabular}}} &
  \cellcolor[HTML]{E2E1E1}+Ours &
  \cellcolor[HTML]{E2E1E1}58.1 &
  \cellcolor[HTML]{E2E1E1}60.4 &
  \cellcolor[HTML]{E2E1E1}54.5 &
  \cellcolor[HTML]{E2E1E1}58.9 &
  \cellcolor[HTML]{E2E1E1}47.4 &
  \cellcolor[HTML]{E2E1E1}47.8 &
  \cellcolor[HTML]{E2E1E1}48.6 &
  \cellcolor[HTML]{E2E1E1}49.1 &
  \cellcolor[HTML]{E2E1E1}53.1 \\ \hline
\end{tabular}%
}
\vspace{-0.5em}
\caption{Impact on image and video-based VLMs on the original NExT-QA, IntentQA, and VideoMME test sets. Our framework increases accuracy of all video- and image-based VLMs by 1-4\% on average across all data partitions. Temporal and action partitions benefit most.}
\label{tab:benchmark}
\end{table*}
\begin{table*}[t]
\centering
\resizebox{\textwidth}{!}{%
\begin{tabular}{cc|cc|cc|cc|cc|cc}
\hline
\multicolumn{2}{c|}{Model} &
  BLIP-2 &
  +Ours &
  LLaVA &
  +Ours &
  \begin{tabular}[c]{@{}c@{}}Video\\ Chat2\end{tabular} &
  +Ours &
  \begin{tabular}[c]{@{}c@{}}Video\\ LLaVA\end{tabular} &
  +Ours &
  \begin{tabular}[c]{@{}c@{}}Video\\ LLaMA\end{tabular} &
  +Ours \\ \hline
 & Original & 37.2 & 43.6 & 39.3 & 46.8 & 59.5 & 59.7 & 58.2 & 60.5 & 57.8 & 59.3 \\
\multirow{-2}{*}{NExT-QA} &
  \cellcolor[HTML]{E2E1E1}Rewritten &
  \cellcolor[HTML]{E2E1E1}33.5 &
  \cellcolor[HTML]{E2E1E1}39.8 &
  \cellcolor[HTML]{E2E1E1}34.8 &
  \cellcolor[HTML]{E2E1E1}44.9 &
  \cellcolor[HTML]{E2E1E1}45.4 &
  \cellcolor[HTML]{E2E1E1}49.0 &
  \cellcolor[HTML]{E2E1E1}51.1 &
  \cellcolor[HTML]{E2E1E1}55.4 &
  \cellcolor[HTML]{E2E1E1}41.4 &
  \cellcolor[HTML]{E2E1E1}47.0 \\ \hline
 & Original & 46.2 & 50.7 & 47.9 & 51.6 & 61.8 & 63.1 & 57.3 & 59.7 & 55.9 & 56.7 \\
\multirow{-2}{*}{Intent-QA} &
  \cellcolor[HTML]{E2E1E1}Rewritten &
  \cellcolor[HTML]{E2E1E1}38.2 &
  \cellcolor[HTML]{E2E1E1}45.5 &
  \cellcolor[HTML]{E2E1E1}42.7 &
  \cellcolor[HTML]{E2E1E1}48.6 &
  \cellcolor[HTML]{E2E1E1}52.6 &
  \cellcolor[HTML]{E2E1E1}55.7 &
  \cellcolor[HTML]{E2E1E1}50.5 &
  \cellcolor[HTML]{E2E1E1}54.7 &
  \cellcolor[HTML]{E2E1E1}46.3 &
  \cellcolor[HTML]{E2E1E1}50.0 \\ \hline
\multicolumn{2}{c|}{Avg} &
  38.8 &
  44.9 &
  41.2 &
  48.0 &
  54.8 &
  56.9 &
  54.3 &
  57.6 &
  50.4 &
  53.3 \\ \hline
\end{tabular}%
}
    \vspace{-0.5em}
\caption{Results on de-biased QA sets. Video-based VLMs show significant decreases in the rewritten de-biased set. In contrast, our framework demonstrates much greater robustness on the rewritten set.}
\label{tab:de-bias}
\end{table*}

\noindent \textbf{Datasets.}
We test our framework on three VQA benchmarks: (1) \textit{NExT-QA}~\cite{nextqa}, a VQA benchmark for causal and temporal reasoning. (2) \textit{IntentQA}~\cite{intentqa}, which focuses on video intent reasoning from both causal and temporal aspects.  (3) \textit{Video-MME}~\cite{videomme}, which is a recently proposed comprehensive evaluation benchmark for video analysis; we use 
its ``short-term'' split (video length $<$ 2 mins) and 4 question types (temporal, spatial, action, and object reasoning) highly related to commonsense reasoning are selected.

\noindent \textbf{Evaluation.}
We report model performances on our rewritten test set for each dataset and its original test set. 
We evaluate our framework on all datasets under the multiple-choice QA setting, using a standard accuracy metric. 

\noindent \textbf{Baselines.} Our baselines represent three categories:
\begin{itemize}
    \item \textit{Video-based VLMs:} Video-based VLMs are widely used for VQA tasks, so we include Video-LLaVA~\cite{videollava}, VideoChat2~\cite{videochat2}, and VideoLLaMA~\cite{videollama}. To test the effectiveness of our framework, we integrate these VLMs by replacing our  \texttt{Prv}${(\cdot)}$ with specific VLM models. 
    \item \textit{Image-based VLMs:} 
    We include BLIP-2~\cite{blip2} and LLaVA-1.5~\cite{llava} as Image-LLM baselines.
    \item \textit{State-of-the-art VQA approaches:} Recent works in VQA, such as VideoTree~\cite{videotree}, VideoAgent~\cite{videoagent}, and LLoVi~\cite{llovi}, are included as strong baselines.
\end{itemize}

\noindent \textbf{Implementation details.} 
Our entire framework is training- and human annotation-free. We use LLaMA-3-8B~\cite{llama} to handle basic functionalities, including (1) converting original QA into declarative statements, (2) statement decomposition, (3) structured semantic extraction and retrieval, and (4) guiding evidence grounding. Detailed prompts for each functionality are provided in the supplementary material. For frame-wise captioning across all datasets, we use LLaVA-1.5~\cite{llava} as our default captioner. When comparing with state-of-the-art VQA methods (e.g., VideoAgent), we follow their setup by replacing the captioner with the stronger CogAgent~\cite{cogagent} model for fair comparison. When integrating our framework with VLMs, the VLM itself serves as the Prover, \texttt{Prv}$(\cdot)$. For video-LLMs, frames are uniformly sampled from the grounded video moment to meet their input requirements (VideoChat2: 16 frames; VideoLLaVA \& VideoLLaMA: 8 frames). For image-language models like LLaVA, we sample 8 frames from the grounded moment and process them individually; the final confidence score is obtained by averaging scores across frames. During dynamic tree generation, we also set a max depth of 5 for the overall entailment tree to improve the efficiency.

\begin{table*}[t]
\centering
\resizebox{\textwidth}{!}{%
\begin{tabular}{l|c|cc|cc|cccc}
\hline
Method     & Model & \multicolumn{2}{c|}{NExT-QA*} & \multicolumn{2}{c|}{IntentQA*} & \multicolumn{4}{c}{VideoMME}                \\ \cline{3-10} 
             & Reasoner      & Temporal         & Causal        & Temporal         & Causal         & Temporal   & Spatial   & Action            & Object            \\ \hline
VideoAgent~\cite{videoagent}            & GPT-4 (1.8T)  & 58.2          & \textbf{66.6}         & 60.4          & 61.0          & -    & -    & -             & -             \\
VideoTree~\cite{videotree}             & GPT-4 (1.8T)  & 60.2          & 66.4         & 56.7          & 60.1          & 55.7 & 54.3 & \textbf{54.2} & \textbf{52.6} \\
LLoVi~\cite{llovi}
   &
  GPT-4 (1.8T) &
  \cellcolor[HTML]{FFFFFF}53.1 &
  \cellcolor[HTML]{FFFFFF}60.8 &
  \cellcolor[HTML]{FFFFFF}58.6 &
  \cellcolor[HTML]{FFFFFF}61.8 &
  \cellcolor[HTML]{FFFFFF}52.2 &
  \cellcolor[HTML]{FFFFFF}\textbf{55.3} &
  \cellcolor[HTML]{FFFFFF}51.8 &
  \cellcolor[HTML]{FFFFFF}50.8 \\
Ours &
  VideoLLAVA (7B) &
  \cellcolor[HTML]{E2E1E1}\textbf{60.8} &
  \cellcolor[HTML]{E2E1E1}65.9 &
  \cellcolor[HTML]{E2E1E1}\textbf{61.0} &
  \cellcolor[HTML]{E2E1E1}\textbf{62.6} &
  \cellcolor[HTML]{E2E1E1}\textbf{55.9} &
  \cellcolor[HTML]{E2E1E1}53.8 &
  \cellcolor[HTML]{E2E1E1}54.0 &
  \cellcolor[HTML]{E2E1E1}50.8 \\ \hline
\end{tabular}%
}
 \vspace{-0.5em}
\caption{Comparison with state-of-the-art. Results for NExT-QA and IntentQA are reported under the de-biased set (the results on the original sets are similar; we provide them in the Appendix). The `Reasoner'' in these approaches is similar to the ``Prover'' in our framework. The captioner for all methods is CogAgent~\cite{cogagent}. Despite other methods relying on much stronger reasoning models, our approach yields competitive performance (four state-of-the-art results) and high parameter efficiency (\textbf{257× fewer} than GPT-4 reasoners).}
\label{tab:sota_qa}
\end{table*}
\subsection{Main results}
\vspace{-0.5em}

\noindent \textbf{Benefit for 
image and video-based VLMs.} \cref{tab:benchmark} summarizes the results of our method compared to baselines. Integrating entailment tree reasoning brings consistent improvement across the video- and image-based VLMs for all datasets (1-4\% on average). 
This finding includes the recently proposed benchmark VideoMME, which poses much more challenging videos and questions. 
The benefits are particularly regular for temporal reasoning (improvement in 14 out of 15 cases), which illustrates how our explicit reasoning process enhances temporal commonsense QA. Image-based VLMs, which initially lack temporal modeling capabilities, perform poorly when directly applied to video QA tasks. However, our framework provides a significant performance boost for these models up to 8\% for LLAVA-1.5. By reasoning over multiple sub-problems rather than tackling the entire complex question simultaneously, our reasoning method makes the task more manageable for both video- and image-based VLMs.

\noindent \textbf{Results on de-biased QA sets.} As shown in \cref{tab:de-bias}, all models experience considerable performance drops on the de-biased set of the same VQA dataset, which aligns with our observation that current VLMs often rely on textual bias in commonsense reasoning tasks. Notably, video-based VLMs show an 8\%-10\% decrease in the de-biased set even though the question and the correct answer remain unchanged. In contrast, our proposed framework, which derives answers through an explicit reasoning process based on specific visual evidence, demonstrates much greater robustness on the de-biased set. The improvement brought by our framework on the de-biased set is even higher than
on the original test sets. In turn, our framework compensates for the performance loss of the VLMs on the de-biased set.
This analysis underscores our framework's potential to mitigate textual bias in commonsense reasoning. Furthermore, the performance differences between the original and de-biased QA sets highlight VQA benchmarks' limitations in evaluating VLMs' true reasoning abilities.

\noindent \textbf{Comparison with state-of-the-art.} 
\cref{tab:sota_qa} compares our framework's results on the de-biased sets to state-of-the-art VQA approaches. The table shows that, next to the consistent benefit our framework provides to various VLMs, it is also competitive with state-of-the-art VQA methods.  When applying an advanced captioner and reasoner that aligns with VideoAgent and VideoTree, our framework yields new state-of-the-art results in some cases. In particular, our framework performs best on temporal reasoning questions for all three benchmarks and outperforms all methods on the IntentQA dataset. Importantly, our method reaches such competitive performance despite using \textbf{$257\times$ fewer} parameters for its reasoning compared to state-of-the-art methods.

\subsection{Ablation studies}
\vspace{-0.5em}
Ablation experiments are conducted using VideoLLaVA's baseline with our framework. Results are reported on the test set of the NExT-QA dataset. 

\begin{table}[t]
\centering
\begin{tabular}{l l|cc}
\hline
\multirow{2}{*}{Model class} & \multirow{2}{*}{LLM} & \multicolumn{2}{c}{NExT-QA} \\ \cline{3-4} 
                  &    & Original        & Rewritten       \\ \hline
\multirow{3}{*}{Open-source} & Mistral-7B           & 60.0            & 55.6            \\
 & LLaMA-3-8B           & 60.5            & 55.4            \\
& LLaMA-3-70B          & 61.3            & 55.9            \\
 \hline
\multirow{2}{*}{Proprietary} & Gemini-1.5-Pro       & 61.1            & 55.2            \\ 
& GPT-4                & 61.6   & 56.1   \\ \hline
\end{tabular}%
\vspace{-0.5em}
\caption{Impact of LLMs for statement decomposition. The open-source and proprietary models are ordered ascendingly by size. Larger models, especially GPT-4, are best at decomposition, but the smaller models (e.g., LlaMa-3-8B) come close.}
\label{tab:LLMs}
\end{table}
\noindent \textbf{Impact of LLMs for statement decomposition.} Entailment generation in our framework relies on prompting an external LLM to recursively decompose statements (cf. \cref{sec:tree}), which is crucial in guiding reasoning paths. Consequently, we tested various LLMs for entailment tree generation, including open-source models (LLama-3 and Mistral) of different sizes and proprietary LLMs (GPT-4 and Gemini-1.5). The results are summarized in \cref{tab:LLMs}. As expected, the proprietary model GPT-4, known for its strong step-by-step reasoning capabilities, delivers the best performance across all settings. Scaling up LLaMA-3 to 70B offers improvements over the 8B model, though with a notable increase in inference time. As the overall performance difference between models is within 1\%, we select LLaMA-3-8B as the default for integrating our framework into VLMs due to its free availability and efficiency.

\begin{table}[t]
\centering
\resizebox{\linewidth}{!}{
\begin{tabular}{cc|cc}
\hline
\multicolumn{2}{c|}{Components} &
 \multicolumn{2}{c}{\begin{tabular}[c]{@{}c@{}}NExT-QA\end{tabular}} \\ \hline
\begin{tabular}[c]{@{}c@{}}Fact-conditioned\\ captioning\end{tabular} &
  \begin{tabular}[c]{@{}c@{}}Structure-based\\ retrieval\end{tabular} &
  Original &
  Rewritten \\ \hline
  &     & 56.2 & 49.6 \\
$\checkmark$ &     & 59.5 & 53.3 \\
    & $\checkmark$ & 58.4 & 52.7 \\
$\checkmark$ & $\checkmark$ & 60.5 & 55.4 \\ \hline
\end{tabular}}
\vspace{-0.5em}
\caption{Ablation on grounding components, showing that both fact-conditional captioning and structure-guided retrieval enhance overall performance by improving grounding accuracy.} 
\label{tab:break_ground}
\end{table}
\noindent \textbf{Ablation on grounding components.} Next, we test the effectiveness of each component in our grounding module (\cref{sec:verification}). The results, summarized in~\cref{tab:break_ground}, indicate that both fact-conditional captioning and structure-guided retrieval enhance overall performance by improving grounding accuracy. However, using only structure-guided retrieval results in a slight performance drop, possibly because \texttt{Cap}($\cdot$) introduces irrelevant semantic information that doesn’t align with the question’s focus, and the structured representation can make identifying anchor frames more challenging. In contrast, fact-conditional captioning alone yields substantial improvement, demonstrating that this straightforward approach can yield an effective and more controllable textual description for videos by conditioning on prior knowledge or relevant facts.

\begin{table}[t]
\centering
\begin{tabular}{c|ccccc}
\hline
\multirow{2}{*}{\begin{tabular}[c]{@{}c@{}}Acc\\ (NExT-QA)\end{tabular}} & \multicolumn{5}{c}{Frame number} \\ \cline{2-6} 
          & 4    & 8    & 16   & 24   & 32   \\ \hline
Original  & 57.7 & 59.3 & 60.5 & 60.7 & 61.0 \\
Rewritten & 51.9 & 53.4 & 55.4 & 55.4 & 55.7 \\ \hline
\end{tabular}%
    \vspace{-0.5em}
\caption{Impact of video frame amount. Strong performance requires a sufficiently high frame number (over 16 for NExT-QA).}
\label{tab:length}
\end{table}
\noindent \textbf{Impact of length of video frames.} 
We further ablate the impact of input video frame length in our framework to determine the optimal number of frame-wise captions to generate per video. The results, summarized in~\cref{tab:length}, show that ideal performance is achieved only when sampling a sufficient number of frames (at least 16 for NExT-QA). When fewer frames are used (e.g., 4 or 8), key anchor frames may be missed, reducing the accuracy of grounded visual evidence. Additionally, while increasing the frame count to 32 yields the best performance, it also increases the calls required for \texttt{Cap}$(\cdot)$ to generate frame-wise captions. Balancing efficiency with performance gains, we set 24 frames as the default in our implementation.

\noindent \textbf{Effectiveness of evidence grounding.} Our method grounds relevant video fragments to support statements in the entailment tree (\cref{sec:verification}). To validate its effectiveness, we compare it with two other variations as sources of visual evidence: (1) without evidence grounding, using the full video as evidence, and (2) upper-bound results: manually annotated temporal boundaries provided in the NExT-GQA dataset, indicating where the QA models should focus when producing correct answers. The results are shown in \cref{tab:ground}. Compared to the baseline, our video-grounded method provides consistent improvements across the original and de-biased sets. The improvement is more apparent in the de-biased set, where the answer options are more semantically similar and require more precise, discriminative visual evidence. Using the ground-truth fragment can further boost our approach, suggesting that enhancing grounding accuracy could further improve our framework.

\begin{table}[]
\centering
\resizebox{\linewidth}{!}{%
\begin{tabular}{cc|ccc}
\hline
\multicolumn{2}{c|}{\begin{tabular}[c]{@{}c@{}}Video fragment\end{tabular}} &
  \begin{tabular}[c]{@{}c@{}}Full\end{tabular} &
  \begin{tabular}[c]{@{}c@{}}Grounded (ours)\end{tabular} &
  \begin{tabular}[c]{@{}c@{}}GT\end{tabular} \\ \hline
\multirow{2}{*}{NExT-QA} &
  Original &
  58.3 &
  60.5 &
  61.8 \\ \cline{2-2}
 &
  Rewritten &
  51.7 &
  55.4 &
  56.9 \\ \hline
\end{tabular}%
}
\vspace{-0.5em}
\caption{Effectiveness of evidence grounding. Our video-grounded method yields clear improvement over using the full video. More precise grounding can further enhance our accuracy.}
\label{tab:ground}
\end{table}
\begin{table}[]
\centering
\resizebox{\linewidth}{!}{%
\begin{tabular}{cc|cccl|c}
\hline
\multicolumn{2}{c|}{\multirow{2}{*}{Strategy}} & \multicolumn{4}{c|}{\begin{tabular}[c]{@{}c@{}}Static (Depth=)\end{tabular}} & \multirow{2}{*}{Dynamic} \\ \cline{3-6}
\multicolumn{2}{c|}{}                & 2    & 3    & 4    & 5    &      \\ \hline
\multirow{2}{*}{NExT-QA} & Original  & 58.8 & 59.2 & 60.2 & 60.3 & 60.5 \\ \cline{2-2}
                         & Rewritten & 52.0 & 53.4 & 55.6 & 55.3 & 55.4 \\ \hline
\end{tabular}%
}
\vspace{-0.5em}
\caption{Effectiveness of dynamic tree expansion. It yields superior accuracy while increasing reasoning efficiency.}
\label{tab:dynamic}
\end{table}
\noindent \textbf{Effectiveness of dynamic tree expansion.} The depth of the entailment tree determines the granularity of reasoning (\cref{sec:dynamic}). This ablation analyzes how tree depth impacts overall performance and compares a fixed-depth approach with our dynamic tree generation strategy. Increasing the depth of reasoning yields significant improvements, as complex, long statements are broken down into concise sub-statements that VLMs can understand more effectively. However, extending reasoning beyond the 4th layer offers diminishing returns; for NExT-QA, the original statements' complexity constrains the task, and some 5th-layer sub-statements become overly simplistic and less effective for reasoning. This finding highlights the necessity of our dynamic strategy. Applying the dynamic tree expansion strategy, we can see that the performance outperforms the fixed-depth paradigm. In the meantime, the dynamic strategy increases the reasoning efficiency over the entailment tree, more details about efficiency comparison can be found in our supplementary material.

\section{Conclusion}
\vspace{-0.5em}
This paper proposed the first video-grounded entailment tree framework for VQA. Moreover, we also contributed a de-biasing procedure to avoid spurious correlations during evaluation and applied it to enhance representative benchmarks. Extensive experiments with five video- and image-based VLMs demonstrate consistent benefits of our method on these benchmarks. Besides, our proposed framework performs on par with state-of-the-art video reasoning methods despite using $257\times$ fewer parameters. While de-biasing hurts VLM accuracy, our framework regains the accuracy losses and is competitive with state-of-the-art VQA methods. 
{
    \small
    \bibliographystyle{ieeenat_fullname}
    \bibliography{main}

\begin{thebibliography}{33}
\providecommand{\natexlab}[1]{#1}
\providecommand{\url}[1]{\texttt{#1}}
\expandafter\ifx\csname urlstyle\endcsname\relax
  \providecommand{\doi}[1]{doi: #1}\else
  \providecommand{\doi}{doi: \begingroup \urlstyle{rm}\Url}\fi

\bibitem[Bagad et~al.(2023)Bagad, Tapaswi, and Snoek]{tot}
Piyush Bagad, Makarand Tapaswi, and Cees~GM Snoek.
\newblock Test of time: Instilling video-language models with a sense of time.
\newblock In \emph{Proceedings of the IEEE/CVF Conference on Computer Vision and Pattern Recognition}, pages 2503--2516, 2023.

\bibitem[Besta et~al.(2024)Besta, Blach, Kubicek, Gerstenberger, Podstawski, Gianinazzi, Gajda, Lehmann, Niewiadomski, Nyczyk, et~al.]{graph-of-thought}
Maciej Besta, Nils Blach, Ales Kubicek, Robert Gerstenberger, Michal Podstawski, Lukas Gianinazzi, Joanna Gajda, Tomasz Lehmann, Hubert Niewiadomski, Piotr Nyczyk, et~al.
\newblock Graph of thoughts: Solving elaborate problems with large language models.
\newblock In \emph{Proceedings of the AAAI Conference on Artificial Intelligence}, pages 17682--17690, 2024.

\bibitem[Chen et~al.(2024)Chen, Liu, He, Chen, Gan, Ma, Zhong, Zhang, Wang, Lin, et~al.]{mecd2024}
Tieyuan Chen, Huabin Liu, Tianyao He, Yihang Chen, Chaofan Gan, Xiao Ma, Cheng Zhong, Yang Zhang, Yingxue Wang, Hui Lin, et~al.
\newblock Mecd: Unlocking multi-event causal discovery in video reasoning.
\newblock \emph{arXiv preprint arXiv:2409.17647}, 2024.

\bibitem[Dalvi et~al.(2021)Dalvi, Jansen, Tafjord, Xie, Smith, Pipatanangkura, and Clark]{entailmentbank}
Bhavana Dalvi, Peter Jansen, Oyvind Tafjord, Zhengnan Xie, Hannah Smith, Leighanna Pipatanangkura, and Peter Clark.
\newblock Explaining answers with entailment trees.
\newblock \emph{arXiv preprint arXiv:2104.08661}, 2021.

\bibitem[Dubey et~al.(2024)Dubey, Jauhri, Pandey, Kadian, Al-Dahle, Letman, Mathur, Schelten, Yang, Fan, et~al.]{llama}
Abhimanyu Dubey, Abhinav Jauhri, Abhinav Pandey, Abhishek Kadian, Ahmad Al-Dahle, Aiesha Letman, Akhil Mathur, Alan Schelten, Amy Yang, Angela Fan, et~al.
\newblock The llama 3 herd of models.
\newblock \emph{arXiv preprint arXiv:2407.21783}, 2024.

\bibitem[Fu et~al.(2024)Fu, Dai, Luo, Li, Ren, Zhang, Wang, Zhou, Shen, Zhang, et~al.]{videomme}
Chaoyou Fu, Yuhan Dai, Yongdong Luo, Lei Li, Shuhuai Ren, Renrui Zhang, Zihan Wang, Chenyu Zhou, Yunhang Shen, Mengdan Zhang, et~al.
\newblock Video-mme: The first-ever comprehensive evaluation benchmark of multi-modal llms in video analysis.
\newblock \emph{arXiv preprint arXiv:2405.21075}, 2024.

\bibitem[Hong et~al.(2024)Hong, Wang, Lv, Xu, Yu, Ji, Wang, Wang, Dong, Ding, et~al.]{cogagent}
Wenyi Hong, Weihan Wang, Qingsong Lv, Jiazheng Xu, Wenmeng Yu, Junhui Ji, Yan Wang, Zihan Wang, Yuxiao Dong, Ming Ding, et~al.
\newblock Cogagent: A visual language model for gui agents.
\newblock In \emph{Proceedings of the IEEE/CVF Conference on Computer Vision and Pattern Recognition}, pages 14281--14290, 2024.

\bibitem[Kassner et~al.(2023)Kassner, Tafjord, Sabharwal, Richardson, Schuetze, and Clark]{rationality}
Nora Kassner, Oyvind Tafjord, Ashish Sabharwal, Kyle Richardson, Hinrich Schuetze, and Peter Clark.
\newblock Language models with rationality.
\newblock \emph{arXiv preprint arXiv:2305.14250}, 2023.

\bibitem[Khattak et~al.(2024)Khattak, Naeem, Hassan, Naseer, Tombari, Khan, and Khan]{benchmark1}
Muhammad~Uzair Khattak, Muhammad~Ferjad Naeem, Jameel Hassan, Muzammal Naseer, Federico Tombari, Fahad~Shahbaz Khan, and Salman Khan.
\newblock Complex video reasoning and robustness evaluation suite for video-lmms.
\newblock \emph{arXiv preprint arXiv:2405.03690}, 2024.

\bibitem[Li et~al.(2023{\natexlab{a}})Li, Li, Savarese, and Hoi]{blip2}
Junnan Li, Dongxu Li, Silvio Savarese, and Steven Hoi.
\newblock Blip-2: Bootstrapping language-image pre-training with frozen image encoders and large language models.
\newblock In \emph{International conference on machine learning}, pages 19730--19742. PMLR, 2023{\natexlab{a}}.

\bibitem[Li et~al.(2023{\natexlab{b}})Li, Wei, Han, and Fan]{intentqa}
Jiapeng Li, Ping Wei, Wenjuan Han, and Lifeng Fan.
\newblock Intentqa: Context-aware video intent reasoning.
\newblock In \emph{Proceedings of the IEEE/CVF International Conference on Computer Vision}, pages 11963--11974, 2023{\natexlab{b}}.

\bibitem[Li et~al.(2024)Li, Wang, He, Li, Wang, Liu, Wang, Xu, Chen, Luo, et~al.]{videochat2}
Kunchang Li, Yali Wang, Yinan He, Yizhuo Li, Yi Wang, Yi Liu, Zun Wang, Jilan Xu, Guo Chen, Ping Luo, et~al.
\newblock Mvbench: A comprehensive multi-modal video understanding benchmark.
\newblock In \emph{Proceedings of the IEEE/CVF Conference on Computer Vision and Pattern Recognition}, pages 22195--22206, 2024.

\bibitem[Li et~al.(2022)Li, Wang, Xiao, Ji, and Chua]{invariant}
Yicong Li, Xiang Wang, Junbin Xiao, Wei Ji, and Tat-Seng Chua.
\newblock Invariant grounding for video question answering.
\newblock In \emph{Proceedings of the IEEE/CVF Conference on Computer Vision and Pattern Recognition}, pages 2928--2937, 2022.

\bibitem[Lin et~al.(2023)Lin, Ye, Zhu, Cui, Ning, Jin, and Yuan]{videollava}
Bin Lin, Yang Ye, Bin Zhu, Jiaxi Cui, Munan Ning, Peng Jin, and Li Yuan.
\newblock Video-llava: Learning united visual representation by alignment before projection.
\newblock \emph{arXiv preprint arXiv:2311.10122}, 2023.

\bibitem[Liu et~al.(2022)Liu, Lv, See, and Lin]{task2022}
Huabin Liu, Weixian Lv, John See, and Weiyao Lin.
\newblock Task-adaptive spatial-temporal video sampler for few-shot action recognition.
\newblock In \emph{Proceedings of the 30th ACM International Conference on Multimedia}, pages 6230--6240, 2022.

\bibitem[Liu et~al.(2023)Liu, Lin, Chen, Li, Li, and See]{VIM2023}
Huabin Liu, Weiyao Lin, Tieyuan Chen, Yuxi Li, Shuyuan Li, and John See.
\newblock Few-shot action recognition via intra-and inter-video information maximization.
\newblock \emph{arXiv preprint arXiv:2305.06114}, 2023.

\bibitem[Liu et~al.(2024)Liu, Li, Li, and Lee]{llava}
Haotian Liu, Chunyuan Li, Yuheng Li, and Yong~Jae Lee.
\newblock Improved baselines with visual instruction tuning.
\newblock In \emph{Proceedings of the IEEE/CVF Conference on Computer Vision and Pattern Recognition}, pages 26296--26306, 2024.

\bibitem[Liu et~al.(2025)Liu, Ma, Zhong, Zhang, and Lin]{timecraft}
Huabin Liu, Xiao Ma, Cheng Zhong, Yang Zhang, and Weiyao Lin.
\newblock Timecraft: Navigate weakly-supervised temporal grounded video question answering via bi-directional reasoning.
\newblock In \emph{European Conference on Computer Vision}, pages 92--107. Springer, 2025.

\bibitem[Sanders et~al.(2024)Sanders, Weir, and Van~Durme]{tvtree}
Kate Sanders, Nathaniel Weir, and Benjamin Van~Durme.
\newblock Tv-trees: Multimodal entailment trees for neuro-symbolic video reasoning.
\newblock \emph{arXiv preprint arXiv:2402.19467}, 2024.

\bibitem[Tafjord et~al.(2022)Tafjord, Mishra, and Clark]{entailer}
Oyvind Tafjord, Bhavana~Dalvi Mishra, and Peter Clark.
\newblock Entailer: Answering questions with faithful and truthful chains of reasoning.
\newblock \emph{arXiv preprint arXiv:2210.12217}, 2022.

\bibitem[Wang et~al.(2024{\natexlab{a}})Wang, Zhang, Zohar, and Yeung-Levy]{videoagent}
Xiaohan Wang, Yuhui Zhang, Orr Zohar, and Serena Yeung-Levy.
\newblock Videoagent: Long-form video understanding with large language model as agent.
\newblock \emph{arXiv preprint arXiv:2403.10517}, 2024{\natexlab{a}}.

\bibitem[Wang et~al.(2024{\natexlab{b}})Wang, Yu, Stengel-Eskin, Yoon, Cheng, Bertasius, and Bansal]{videotree}
Ziyang Wang, Shoubin Yu, Elias Stengel-Eskin, Jaehong Yoon, Feng Cheng, Gedas Bertasius, and Mohit Bansal.
\newblock Videotree: Adaptive tree-based video representation for llm reasoning on long videos.
\newblock \emph{arXiv preprint arXiv:2405.19209}, 2024{\natexlab{b}}.

\bibitem[Wei et~al.(2022)Wei, Wang, Schuurmans, Bosma, Xia, Chi, Le, Zhou, et~al.]{cot}
Jason Wei, Xuezhi Wang, Dale Schuurmans, Maarten Bosma, Fei Xia, Ed Chi, Quoc~V Le, Denny Zhou, et~al.
\newblock Chain-of-thought prompting elicits reasoning in large language models.
\newblock \emph{Advances in neural information processing systems}, 35:\penalty0 24824--24837, 2022.

\bibitem[Weir and Van~Durme(2022)]{nellie}
Nathaniel Weir and Benjamin Van~Durme.
\newblock Dynamic generation of grounded logical explanations in a neuro-symbolic expert system.
\newblock \emph{arXiv preprint arXiv:2209.07662}, 2022.

\bibitem[Wu et~al.(2024)Wu, Liu, Qiao, and Sun]{dibs2024}
Hao Wu, Huabin Liu, Yu Qiao, and Xiao Sun.
\newblock Dibs: Enhancing dense video captioning with unlabeled videos via pseudo boundary enrichment and online refinement.
\newblock In \emph{Proceedings of the IEEE/CVF conference on computer vision and pattern recognition}, pages 18699--18708, 2024.

\bibitem[Xiao et~al.(2021)Xiao, Shang, Yao, and Chua]{nextqa}
Junbin Xiao, Xindi Shang, Angela Yao, and Tat-Seng Chua.
\newblock Next-qa: Next phase of question-answering to explaining temporal actions.
\newblock In \emph{Proceedings of the IEEE/CVF conference on computer vision and pattern recognition}, pages 9777--9786, 2021.

\bibitem[Xiao et~al.(2024)Xiao, Yao, Li, and Chua]{nextgqa}
Junbin Xiao, Angela Yao, Yicong Li, and Tat-Seng Chua.
\newblock Can i trust your answer? visually grounded video question answering.
\newblock In \emph{Proceedings of the IEEE/CVF Conference on Computer Vision and Pattern Recognition}, pages 13204--13214, 2024.

\bibitem[Yang et~al.(2023)Yang, Nushi, Palangi, and Mirzasoleiman]{yang2023mitigating}
Yu Yang, Besmira Nushi, Hamid Palangi, and Baharan Mirzasoleiman.
\newblock Mitigating spurious correlations in multi-modal models during fine-tuning.
\newblock In \emph{International Conference on Machine Learning}, pages 39365--39379. PMLR, 2023.

\bibitem[Yao et~al.(2024)Yao, Yu, Zhao, Shafran, Griffiths, Cao, and Narasimhan]{tree-of-thought}
Shunyu Yao, Dian Yu, Jeffrey Zhao, Izhak Shafran, Tom Griffiths, Yuan Cao, and Karthik Narasimhan.
\newblock Tree of thoughts: Deliberate problem solving with large language models.
\newblock \emph{Advances in Neural Information Processing Systems}, 36, 2024.

\bibitem[Yu et~al.(2024)Yu, Cho, Yadav, and Bansal]{seviLA}
Shoubin Yu, Jaemin Cho, Prateek Yadav, and Mohit Bansal.
\newblock Self-chained image-language model for video localization and question answering.
\newblock \emph{Advances in Neural Information Processing Systems}, 36, 2024.

\bibitem[Zhang et~al.(2023{\natexlab{a}})Zhang, Lu, Islam, Wang, Yu, Bansal, and Bertasius]{llovi}
Ce Zhang, Taixi Lu, Md~Mohaiminul Islam, Ziyang Wang, Shoubin Yu, Mohit Bansal, and Gedas Bertasius.
\newblock A simple llm framework for long-range video question-answering.
\newblock \emph{arXiv preprint arXiv:2312.17235}, 2023{\natexlab{a}}.

\bibitem[Zhang et~al.(2024)Zhang, Zhang, Zhang, and Tresp]{zhang2024can}
Gengyuan Zhang, Yurui Zhang, Kerui Zhang, and Volker Tresp.
\newblock Can vision-language models be a good guesser? exploring vlms for times and location reasoning.
\newblock In \emph{Proceedings of the IEEE/CVF Winter Conference on Applications of Computer Vision}, pages 636--645, 2024.

\bibitem[Zhang et~al.(2023{\natexlab{b}})Zhang, Li, and Bing]{videollama}
Hang Zhang, Xin Li, and Lidong Bing.
\newblock Video-llama: An instruction-tuned audio-visual language model for video understanding.
\newblock \emph{arXiv preprint arXiv:2306.02858}, 2023{\natexlab{b}}.

\end{thebibliography}
}
\clearpage
\setcounter{page}{1}
\maketitlesupplementary

\section{Additional quantitative results}
\noindent \textbf{Comparison with state-of-the-art.} In addition to the state-of-the-art comparison of VQA methods on the de-biased set (\cref{tab:sota_qa} in the main manuscript), we also provide the comparison results on the original test set. Our framework remains competitive with state-of-the-art VQA methods, though our reasoner is about \textbf{$\mathbf{250\times}$ smaller} in parameters than other methods. Moreover, it also achieves new state-of-the-art results in some cases, especially for temporal reasoning. We also notice that the superiority of our framework in the de-biased sets is more significant than in the original sets. This observation highlights the effectiveness of our framework in reasoning over joint visual-text information when the reliance on textual biases is mitigated.

\noindent \textbf{Results on Env-QA.} To further validate the effectiveness and generalizability of our framework, we also test on the Env-QA dataset~\mycite{[1]}, which mainly consists of ego-centric videos collected from virtual environments. We report the results under three types of questions (state, event, and order reasoning), focusing on temporal reasoning. Results are summarized in~\cref{tab:env_qa}. We observe that incorporating our framework brings consistent improvement across the video- and image-based VLMs.

\section{Quality assessment of de-biased set}
We conducted a human evaluation to assess the quality of our de-biased set.  Specifically, we randomly selected 1000 original QA samples and 1000 de-biased QA from the NeXT-QA dataset and presented them to four volunteers.  The volunteers were required to select the best answer from all available options under two distinct conditions: (1) without watching the associated video content, and (2) with the video content available for reference. Results are summarized in~\cref{tab:supp_quality}.
It can be seen that humans can reliably answer rewritten questions (94\%), comparably to the original set (96\%). Meanwhile, in the original set, humans confirm the textual biases and can achieve an accuracy of 79\% without analyzing the video; yet, they cannot easily deduce the correct answer solely from the de-biased question-answer pairs (accuracy of 44\%). 
Hence, our de-biased QA ensures all options pose a comparable level of commonsensical association rather than having a dominant association to the correct answer. It demonstrates that our de-biasing procedure retains the fairness of the benchmark while effectively reducing the textual shortcuts. 

\section{Additional ablation studies}

\noindent \textbf{Design of anchor frame localization.} In our implementation, we directly prompt an LLM to retrieve the anchor frame based on the structured representations of both the fact statement and candidate frames. Additionally, we test other available metrics for anchor frame localization, including (1) \textit{visual-text similarity}: calculating frame-question similarity using CLIP; (2) \textit{text-text similarity}: measuring the similarity between text embeddings of frame-wise captions and the question text; and (3) \textit{LLM-evaluated relevance score}: following the Video-Tree approach~\mycite{[1]}, we prompt the LLM to assign a relevance score to each frame based on its caption and the question text. The comparison results, summarized in~\cref{tab:anchor}, show that our solution performs better than all competitors. Notably, the LLM-evaluated relevance score demonstrates comparable performance to our method, while traditional visual-text and text-text similarity metrics lag behind. This indicates that modern LLMs are highly effective and generalizable tools for approximate retrieval.

\begin{table}[t]
    \centering
    \begin{tabular}{c|c|c}
    \hline
    Method          & Original set & De-biased set \\ \hline
    Human w/o video & 79.3         & 44.6          \\ \hline
    Human w/ video  & 96.4         & 93.9          \\ \hline
    \end{tabular}
    \caption{Results of subjective human evaluation for NeXT-QA, which are derived from the average accuracy of four volunteers.}
    \label{tab:supp_quality}
\end{table}
\begin{table*}[t]
\centering
\small
\begin{tabular}{c|c|cc|cc|cccc}
\hline
Method &
   &
  \multicolumn{2}{c|}{NExT-QA} &
  \multicolumn{2}{c|}{IntentQA} &
  \multicolumn{4}{c}{VideoMME} \\ \cline{3-10} 
 &
  \multirow{-2}{*}{\begin{tabular}[c]{@{}c@{}}Model\\ (Reasoner)\end{tabular}} &
  Temporal &
  Causal &
  Temporal &
  Causal &
  Temporal &
  Spatial &
  Action &
  Object \\ \hline
VideoAgent &
  GPT-4 (1.8T) &
  64.5 &
  72.7 &
  64.1 &
  66.5 &
  - &
  - &
  - &
  - \\
VideoTree &
  GPT-4 (1.8T) &
  \textbf{67.0} &
  \textbf{75.2} &
  61.9 &
  66.1 &
  55.7 &
  54.3 &
  \textbf{54.2} &
  \textbf{52.6} \\
LLoVi &
  GPT-4 (1.8T) &
  \cellcolor[HTML]{FFFFFF}61.0 &
  \cellcolor[HTML]{FFFFFF}69.5 &
  \cellcolor[HTML]{FFFFFF}65.5 &
  \cellcolor[HTML]{FFFFFF}\textbf{68.7} &
  \cellcolor[HTML]{FFFFFF}52.2 &
  \cellcolor[HTML]{FFFFFF}\textbf{55.3} &
  \cellcolor[HTML]{FFFFFF}51.8 &
  \cellcolor[HTML]{FFFFFF}50.8 \\
Ours &
  VideoLLAVA (7B) &
  \cellcolor[HTML]{E2E1E1}64.8 &
  \cellcolor[HTML]{E2E1E1}68.3 &
  \cellcolor[HTML]{E2E1E1}\textbf{66.1} &
  \cellcolor[HTML]{E2E1E1}66.4 &
  \cellcolor[HTML]{E2E1E1}\textbf{55.9} &
  \cellcolor[HTML]{E2E1E1}53.8 &
  \cellcolor[HTML]{E2E1E1}54.0 &
  \cellcolor[HTML]{E2E1E1}50.8 \\ \hline
\end{tabular}%
\caption{Comparison results with state-of-the-art. Results for NExT-QA, IntentQA, and VideoMME are reported under its original test set. The `Reasoner'' in these approaches is similar to the ``Prover'' in our framework. The captioner for all methods is CogAgent. Despite other methods relying on a much stronger reasoning model, our approach yields competitive performance and reaches state-of-the-art results in four out of eight data partitions. Moreover, the reasoner we adopted is $250\times$ smaller than the others.}
\label{tab:sota_qa_orginal}
\end{table*}
\begin{table}[t]
\centering
\resizebox{\linewidth}{!}{%
\begin{tabular}{cc|cccc}
\hline
\multicolumn{2}{c|}{} &
  \multicolumn{4}{c}{Env-QA} \\ \cline{3-6} 
\multicolumn{2}{c|}{\multirow{-2}{*}{Model}} &
  State &
  Event &
  \multicolumn{1}{c|}{Order} &
  Avg \\ \hline
\multicolumn{1}{c|}{} &
  BLIP-2 &
  30.6 &
  28.8 &
  \multicolumn{1}{c|}{40.2} &
  33.2 \\
\multicolumn{1}{c|}{} &
  \cellcolor[HTML]{E2E1E1}+Ours &
  \cellcolor[HTML]{E2E1E1}39.5 &
  \cellcolor[HTML]{E2E1E1}34.5 &
  \multicolumn{1}{c|}{\cellcolor[HTML]{E2E1E1}45.8} &
  \cellcolor[HTML]{E2E1E1}\begin{tabular}[c]{@{}c@{}}39.9 \\ (+6.7)\end{tabular} \\ \cline{2-6} 
\multicolumn{1}{c|}{} &
  LLaVA-1.5 &
  31.3 &
  30.7 &
  \multicolumn{1}{c|}{42.8} &
  34.9 \\
\multicolumn{1}{c|}{\multirow{-4}{*}{\begin{tabular}[c]{@{}c@{}}Image-based\\ VLMs\end{tabular}}} &
  \cellcolor[HTML]{E2E1E1}+Ours &
  \cellcolor[HTML]{E2E1E1}40.5 &
  \cellcolor[HTML]{E2E1E1}36.1 &
  \multicolumn{1}{c|}{\cellcolor[HTML]{E2E1E1}46.2} &
  \cellcolor[HTML]{E2E1E1}\begin{tabular}[c]{@{}c@{}}40.9\\ (+6.0)\end{tabular} \\ \hline
\multicolumn{1}{c|}{} &
  VideoChat2 &
  61.7 &
  49.8 &
  \multicolumn{1}{c|}{60.5} &
  57.3 \\
\multicolumn{1}{c|}{} &
  \cellcolor[HTML]{E2E1E1}+Ours &
  \cellcolor[HTML]{E2E1E1}63.9 &
  \cellcolor[HTML]{E2E1E1}55.1 &
  \multicolumn{1}{c|}{\cellcolor[HTML]{E2E1E1}62.8} &
  \cellcolor[HTML]{E2E1E1}\begin{tabular}[c]{@{}c@{}}60.6\\ (+3.3)\end{tabular} \\ \cline{2-6} 
\multicolumn{1}{c|}{} &
  VideoLLaVA &
  60.5 &
  50.4 &
  \multicolumn{1}{c|}{61.0} &
  57.3 \\
\multicolumn{1}{c|}{\multirow{-4}{*}{\begin{tabular}[c]{@{}c@{}}Video-based\\ VLMs\end{tabular}}} &
  \cellcolor[HTML]{E2E1E1}+Ours &
  \cellcolor[HTML]{E2E1E1}63.3 &
  \cellcolor[HTML]{E2E1E1}55.5 &
  \multicolumn{1}{c|}{\cellcolor[HTML]{E2E1E1}63.2} &
  \cellcolor[HTML]{E2E1E1}\begin{tabular}[c]{@{}c@{}}60.7\\ (+3.4)\end{tabular} \\ \hline
\end{tabular}%
}
\caption{Results on Env-QA. Incorporating our framework brings consistent improvement across the video- and image-based VLMs.}
\label{tab:env_qa}
\end{table}
\begin{table}[t]
\centering
\small
\begin{tabular}{l|c|cc}
\hline
\multirow{2}{*}{Metric}  & \multirow{2}{*}{Model}      & \multicolumn{2}{c}{NExT-QA} \\
& & Original & Rewritten \\ \hline
Visual-text & CLIP & 58.7  & 52.9\\ 
Text-text & LLaMA-3-8B & 58.8 & 52.7\\ 
LLM-score &   LLaMA-3-8B  & 59.7  & 54.3 \\ 
Ours & LLaMA-3-8B  & 60.5 & 55.4 \\ \hline
\end{tabular}%
\caption{Design of anchor frame localization. Our localization LLM outperforms competitive baselines. LLMs overall show a strong ability to retrieve relevant frames.}
\label{tab:anchor}
\end{table}
\begin{table}[t]
\centering
\resizebox{\linewidth}{!}{%
\begin{tabular}{cc|c|cl}
\hline
\multicolumn{2}{c|}{Modality}        & Video-text & \multicolumn{2}{c}{Text} \\ \hline
\multicolumn{2}{c|}{Prv()} & \multicolumn{1}{l|}{VideoLLaVA-7B} & \multicolumn{1}{l}{LLaMA-3-8B} & GPT-4 \\ \hline
\multirow{2}{*}{NExT-QA} & Original  & 60.5       & 57.1        & 59.6       \\ \cline{2-2}
                         & Rewritten & 55.4       & 53.0        & 54.2       \\ \hline
\end{tabular}%
}
\caption{Modality for proving entailment. Text-only reasoning paradigm achieves comparable performance only when a much stronger and larger ($250\times$) LLM, such as GPT-4, is employed.}
\label{tab:prover}
\end{table}
\noindent \textbf{Modality for proving entailment.}
There is a growing trend of transforming multimodal tasks into text-only tasks by converting other modalities into text, enabled by generative multimodal models. This paradigm enables powerful LLMs to tackle challenging tasks more effectively. In our method, we also explore the reasoning paradigm of the prover, comparing our implementation with a purely text-based reasoning solution. Specifically, given captions of the visual evidence for each statement, we directly use an off-the-shelf LLM to assess the confidence score for each statement. The comparison results in~\cref{tab:prover} show that the text-only reasoning paradigm achieves comparable performance when a strong LLM, such as GPT-4, is employed. It is expected that this approach may surpass our method if video-to-text representations are further improved in the future. However, rather than solely focusing on performance, our framework prioritizes providing an interpretable perspective for VLMs in commonsense QA, giving users clear insights into the model’s beliefs and reasoning paths.

\begin{table}[t]
\centering
\small
\begin{tabular}{c|cccc|c}
\hline
\multirow{2}{*}{} & \multicolumn{4}{c|}{Static (Depth=)} & \multirow{2}{*}{Dynamic} \\
                  & 2       & 3       & 4       & 5      &                          \\ \hline
Avg LLM calls     & 1       & 3       & 7       & 15     & 5.6                      \\
Acc (NExT-QA*)              & 52.0    & 53.4    & 55.6    & 55.3   & 55.4                     \\ \hline
\end{tabular}%
\caption{The efficiency comparison between static and dynamic entailment tree generation. `Avg LLM Calls' is the average number of LLM calls made per statement during entailment generation. * indicates the de-biased set. By adopting the dynamic generation strategy, efficiency can be significantly improved without compromising performance.}
\label{tab:eff_dynamic}
\end{table}
\noindent \textbf{Efficiency analysis of dynamic tree generation.} To further validate the necessity of a dynamic strategy in entailment tree generation, we compare the efficiency of static and dynamic entailment tree approaches in~\cref{tab:eff_dynamic}. The results show that the number of LLM calls increases rapidly as the tree depth expands, introducing large time overheads. By adopting the dynamic generation strategy, efficiency can be significantly improved as unnecessary decompositions will be pruned without compromising performance.

\begin{table}[]
\centering
\resizebox{\linewidth}{!}{%
\begin{tabular}{c|cc|cccc}
\hline
\multirow{2}{*}{Method} & \multicolumn{2}{c|}{General VLM} & \multicolumn{4}{c}{VQA approaches}             \\ \cline{2-7} 
                        & VideoChat2      & VideoLlaVA     & VideoAgent & VideoTree & LLoVi & Ours          \\ \hline
Inf time(s)                & 7.5             & 6.2            & 51.0       & 34.6      & 40.3  & 38.2          \\ \hline
Avg acc                 & 49.0            & 50.8           & 61.6       & 60.9      & 58.6  & \textbf{62.6} \\ \hline
Reasoner &
  \begin{tabular}[c]{@{}c@{}}VideoChat2\\ (7B)\end{tabular} &
  \begin{tabular}[c]{@{}c@{}}VideoLlaVA\\ (7B)\end{tabular} &
  \begin{tabular}[c]{@{}c@{}}GPT-4\\ (1.8T)\end{tabular} &
  \begin{tabular}[c]{@{}c@{}}GPT-4\\ (1.8T)\end{tabular} &
  \begin{tabular}[c]{@{}c@{}}GPT-4\\ (1.8T)\end{tabular} &
  \begin{tabular}[c]{@{}c@{}}VideoLlaVA\\ (\textbf{7B})\end{tabular} \\ \hline
\end{tabular}%
}
\caption{Efficiency comparison. The average inference time for each video in the NExT-QA dataset is reported. VideoChat2 and VideoLlaVA are tested using 16 uniformly sampled frames ($224\times 224$) per video. For VideoAgent, VideoTree, and LLoVi, we adhered to their standard post-processing protocols for inference, whereas GPT-4 API served as the reasoning model.}
\label{tab:supp_efficiency}
\end{table}
\noindent \textbf{Efficiency analysis of overall framework} ~\cref{tab:supp_efficiency} presents a comparative analysis of the accuracy-efficiency trade-off between our framework and existing general video-based VLMs, as well as state-of-the-art VQA methods. For this evaluation, we measured the average inference time per video on the NExT-QA dataset using NVIDIA-A600 GPUs. Specifically, VideoChat2 and VideoLlaVA were tested using 16 uniformly sampled frames ($224\times 224$) per video. For VideoAgent, VideoTree, and LLoVi, we adhered to their standard post-processing protocols for inference, whereas GPT-4 API served as the reasoning model. It can be seen that we achieve the best accuracy compared to other methods 
while maintaining a competitive inference speed of 38.2s (faster than 
VideoAgent and LLovi) and high parameter efficiency ($257\times$ fewer 
of the core reasoner 
than GPT-4 reasoners). This parameter efficiency further emphasizes the practicality of our solution.

\section{Qualitative results}
\begin{figure*}[t]
    \centering
    \includegraphics[width=1.0\linewidth]{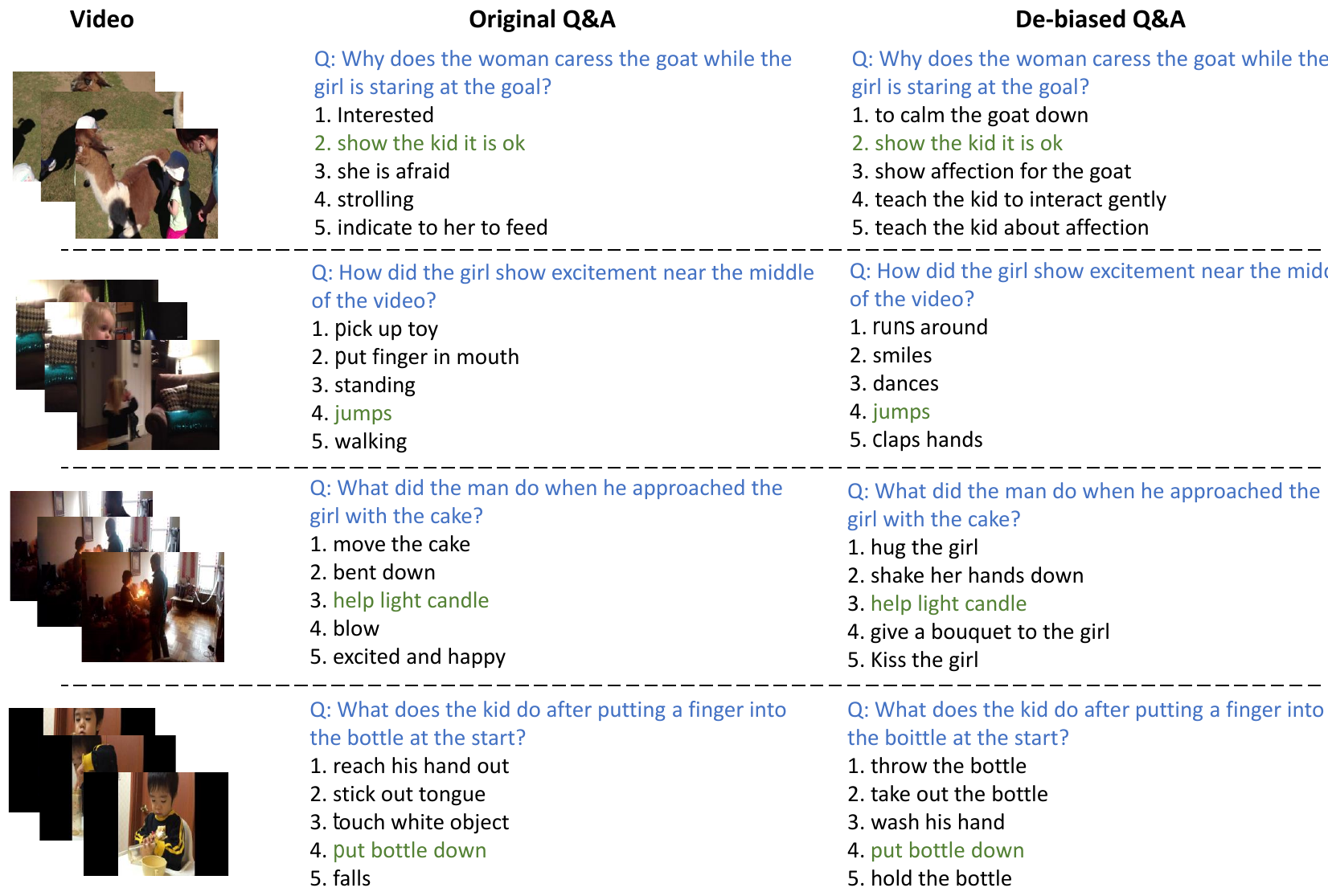}
    \caption{Examples of original and de-biased Q\&A, selected from NExT-QA dataset.}
    \label{fig:example_de_bias}
\end{figure*}

\noindent \textbf{Examples from the de-biased set.} \cref{fig:example_de_bias} showcases examples of Q\&A pairs from the NExT-QA dataset before and after the de-biasing process. The original Q\&A often exhibits textual biases or shortcuts between questions and options, which can be effectively mitigated through answer-set rewriting. The de-biased Q\&A pairs compel VLMs to thoroughly comprehend both the video and text content to arrive at their answers. Therefore, this de-biasing procedure allows a more accurate evaluation of the VLMs' true commonsense reasoning abilities.

\begin{figure*}[t]
    \centering
    \includegraphics[width=1.0\linewidth]{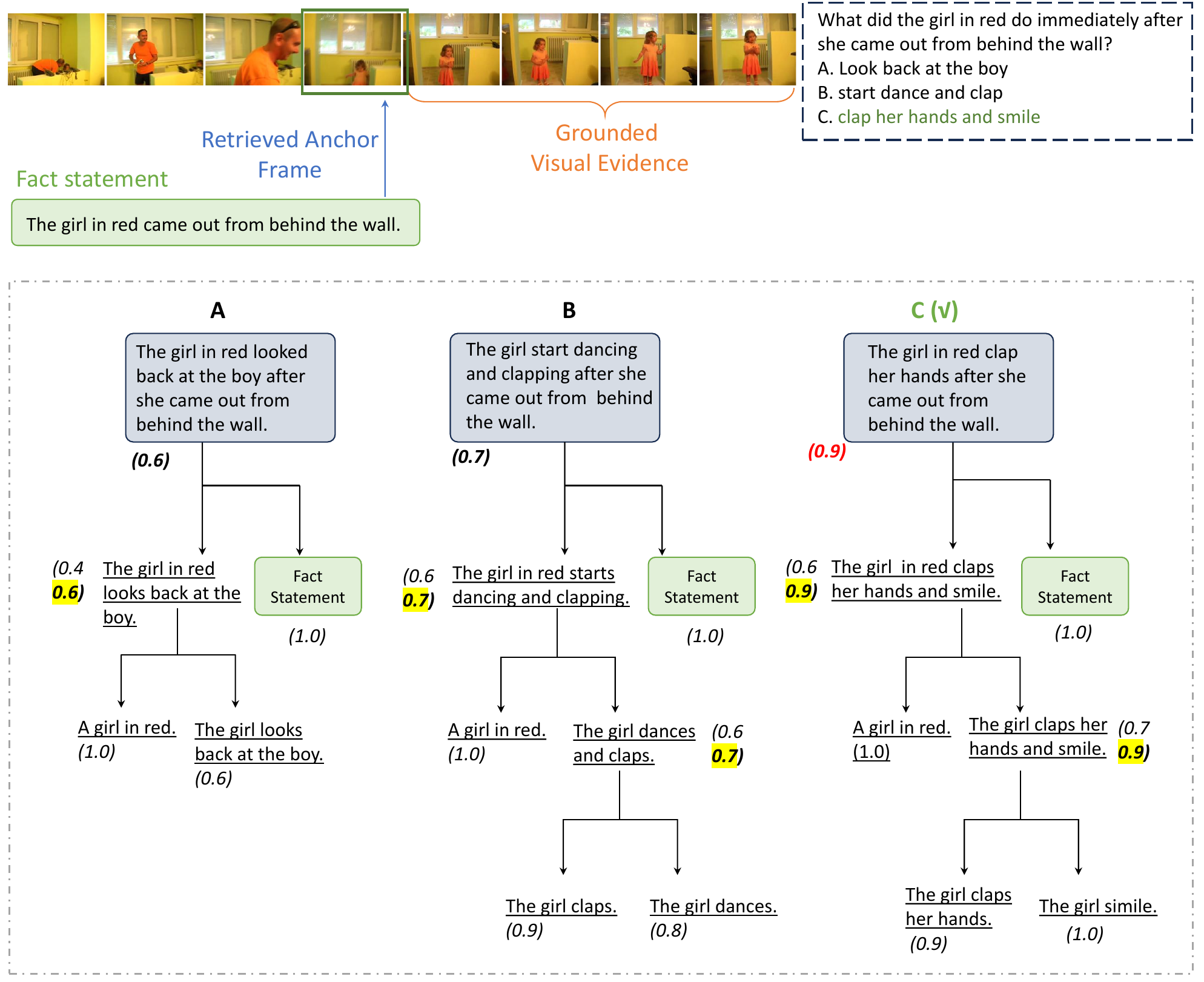}
    \caption{Examples of multi-choice QA inference of our framework. The highlighted confidence score indicates the proof score calculated from child statements.}
    \label{fig:example_qa}
\end{figure*}

\noindent \textbf{Entailment tree reasoning.}
In~\cref{fig:example_qa}, we visualize the Q\&A reasoning process through our proposed framework. Specifically, given the Q\&A pair, we present the entire generated entailment tree and corresponding confidence scores for each statement produced during reasoning. Moreover, the grounded visual evidence is also presented. Our framework provides an interpretable window into VLMs about how the given Q\&A is conducted in both the visual and textual modality.

\section{Additional implementation details} 

\noindent \textbf{Dataset overview} (1) \textbf{NExT-QA} contains 5440 videos with an average length of 44s and approximately 52K questions. NExT-QA contains 3 different question types: Temporal, Causal, and Descriptive. In our experiments, we focus on the commonsense reasoning questions: Temporal and Causal. (2) \textbf{IntentQA} contains 4,303 videos and 16K multiple-choice question-answer pairs focused on reasoning about people's intent in the video. We perform a zero-shot evaluation on the test set containing 2K questions. (3) \textbf{VideoMME} comprises 2,700 QA pairs across 900 videos. Videos are annotated with 12 types of questions, including 4 types specifically designed for commonsense reasoning: temporal reasoning, spatial reasoning, action reasoning, and object reasoning.

\begin{figure*}[t]
    \centering
    \includegraphics[width=1.0\linewidth]{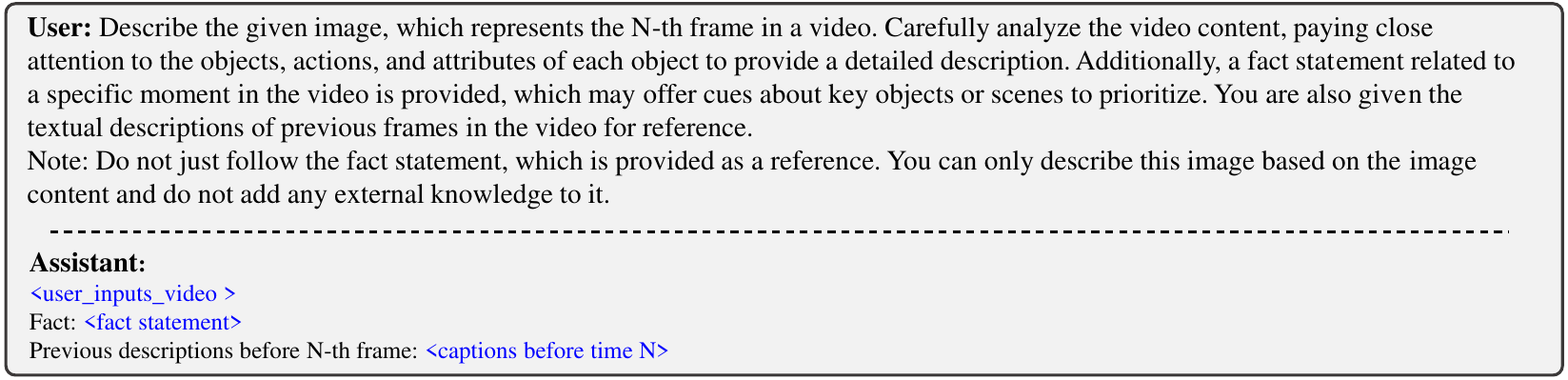}
    \caption{The prompt of fact-conditioned frame captioning for LLaVA-1.5.}
    \label{fig:prompt_captioning}
\end{figure*}

\begin{figure*}[t]
    \centering
    \includegraphics[width=1.0\linewidth]{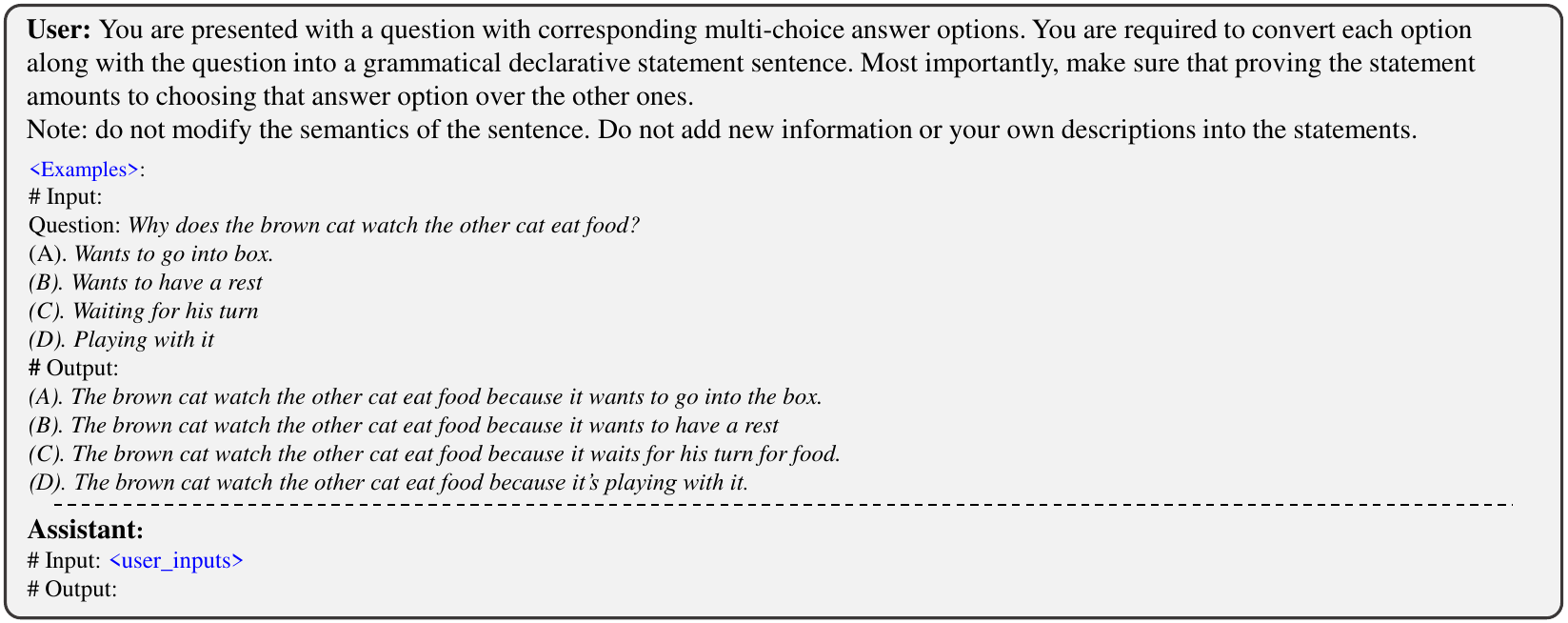}
    \caption{The prompt of transferring Q\&A into declarative statement for LLaMA-3.}
    \label{fig:prompt_declarative}
\end{figure*}

\begin{figure*}[t]
    \centering
    \includegraphics[width=1.0\linewidth]{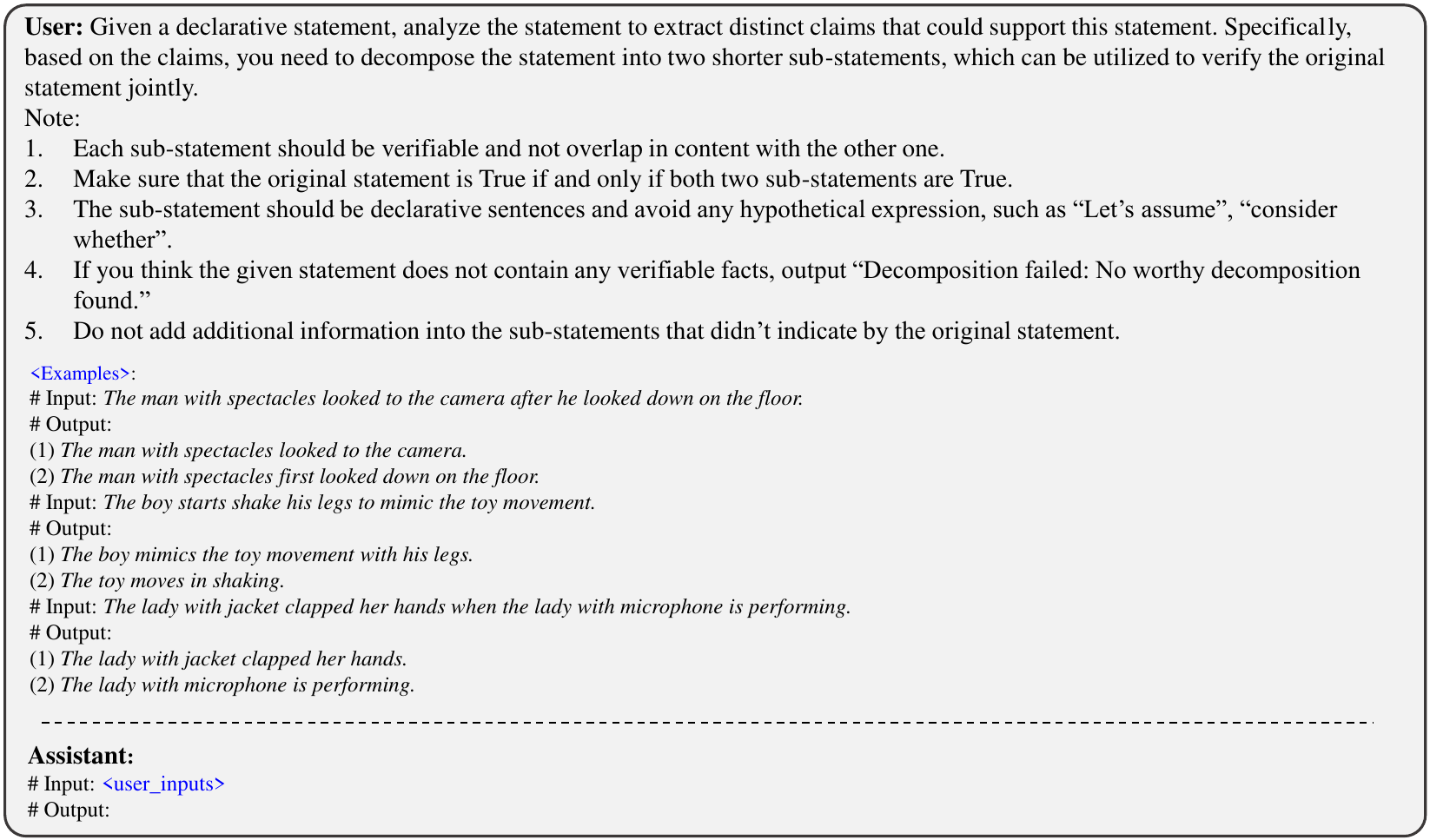}
    \caption{The prompt of statement decomposition for LLaMA-3.}
    \label{fig:prompt_decompose}
\end{figure*}

\begin{figure*}[t]
    \centering
    \includegraphics[width=1.0\linewidth]{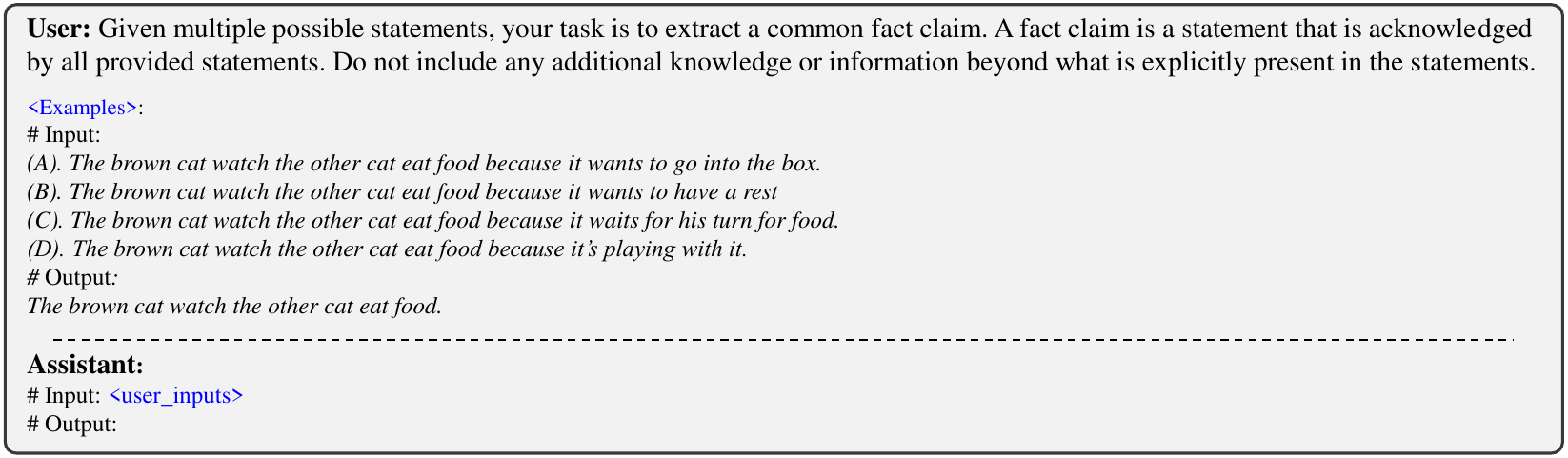}
    \caption{The prompt of fact statement extraction for LLaMA-3.}
    \label{fig:prompt_fact}
\end{figure*}

\begin{figure*}[t]
    \centering
    \includegraphics[width=1.0\linewidth]{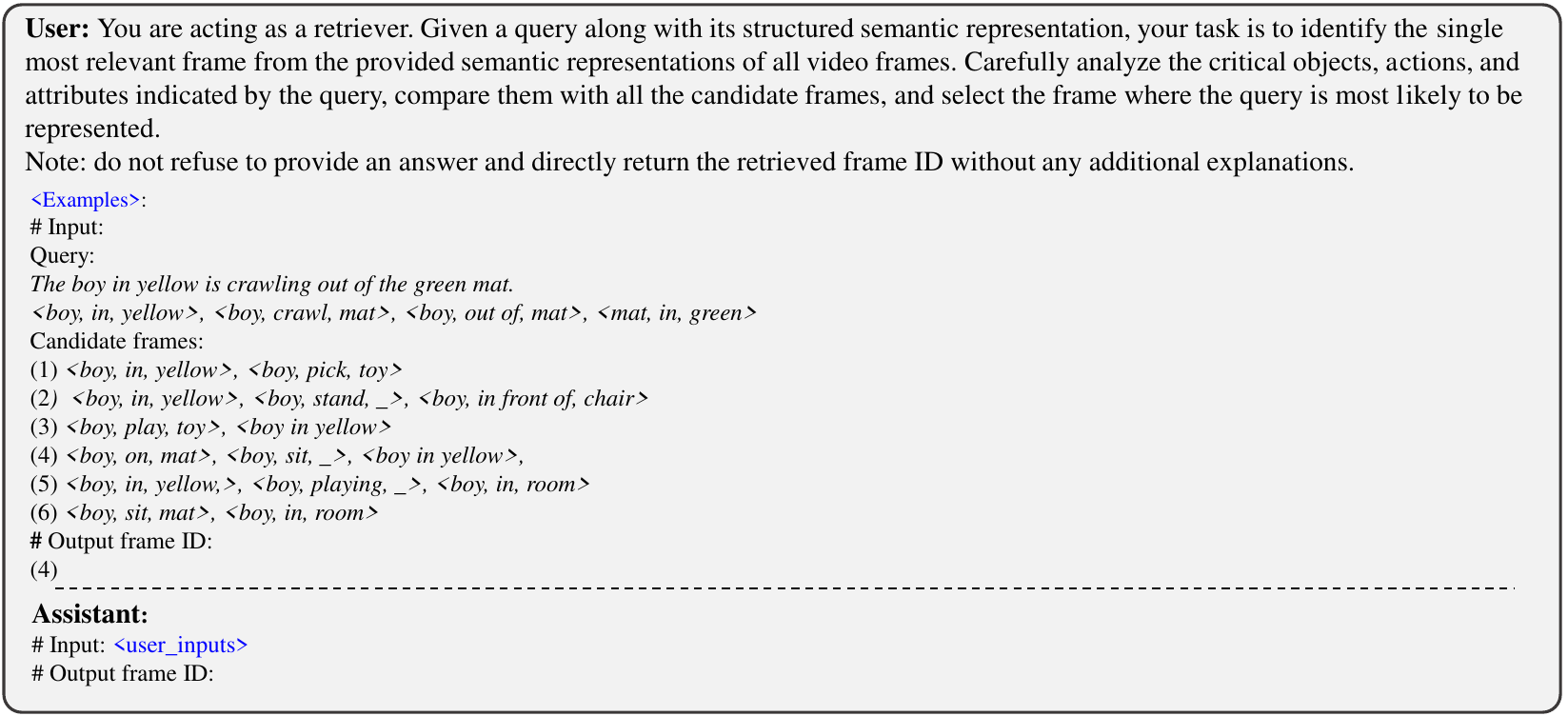}
    \caption{The prompt of retrieving fact statement for LLaMA-3.}
    \label{fig:prompt_retrieval}
\end{figure*}

\begin{figure*}[t]
    \centering
    \includegraphics[width=1.0\linewidth]{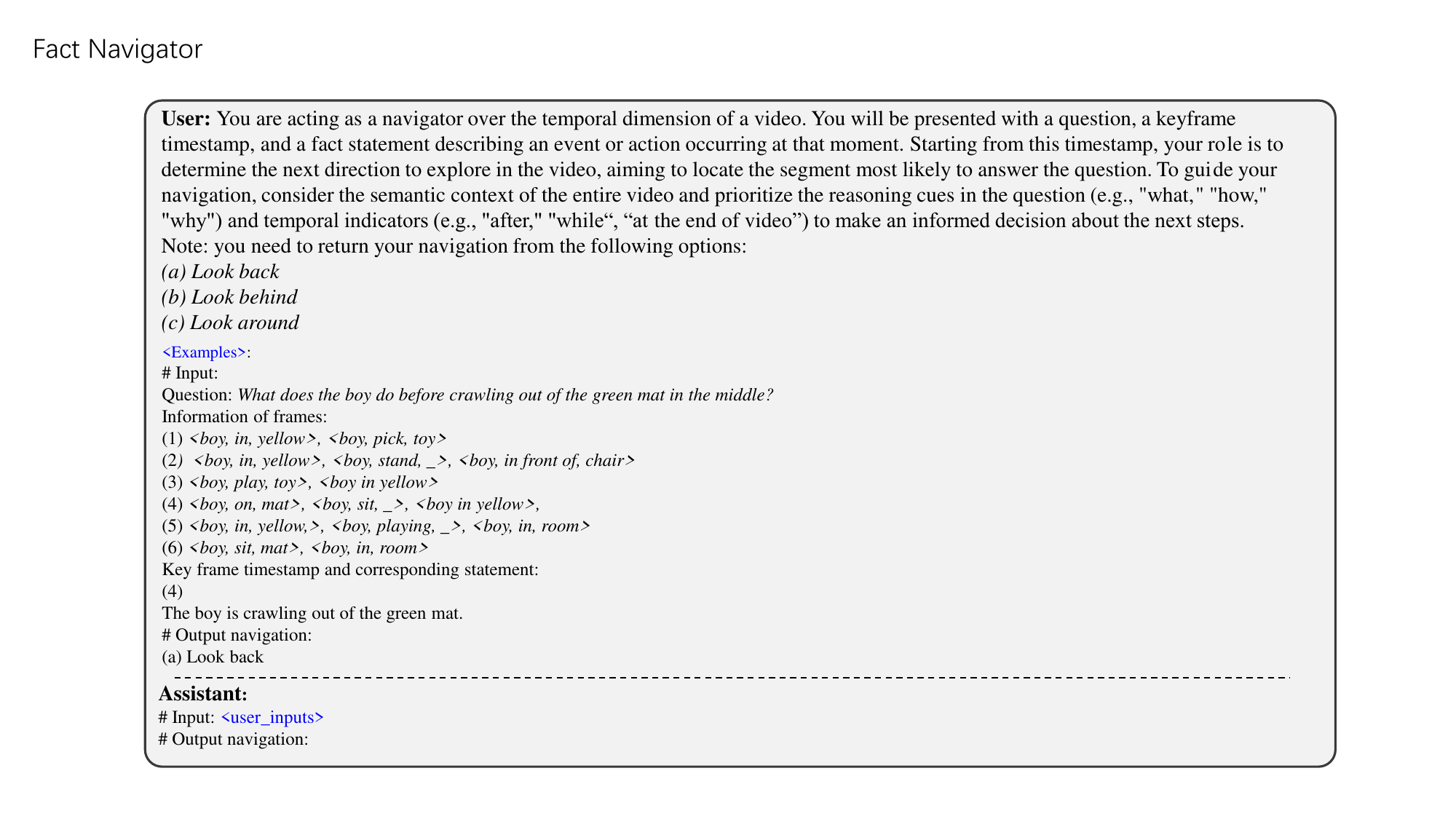}
    \caption{The prompt of evidence navigation for LLaMA-3.}
    \label{fig:prompt_navigator}
\end{figure*}

\begin{figure*}[t]
    \centering
    \includegraphics[width=1.0\linewidth]{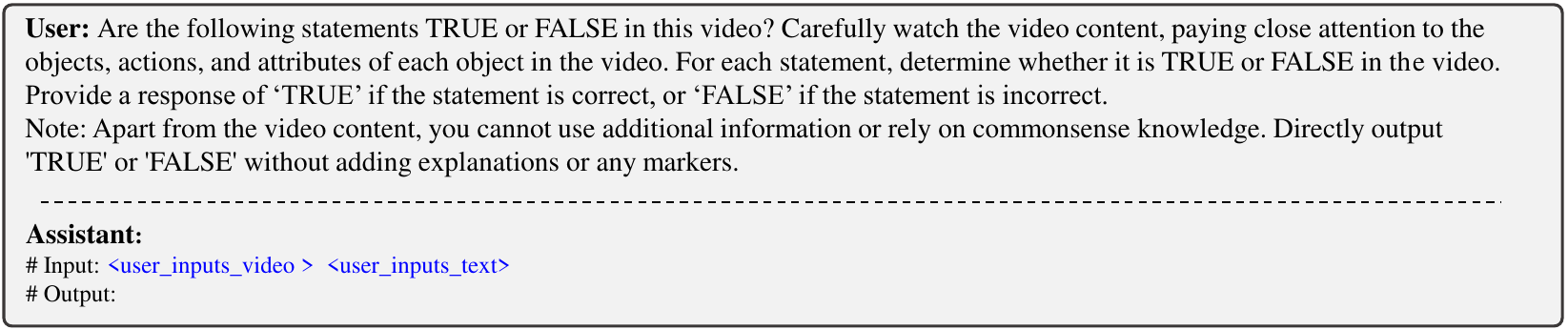}
    \caption{The prompt of statement verification for VideoLLaVA.}
    \label{fig:prompt_verification}
\end{figure*}

\noindent \textbf{Prompt designs.} We provide our detailed designs of LLM prompts for implementing different functionalities in our framework, namely:
\begin{itemize}
    \item \textit{Video captioning}: fact-conditioned frame captioning (\cref{fig:prompt_captioning})
    \item \textit{Entailment tree generation}: declarative statement transformation (\cref{fig:prompt_declarative}), statement decomposition (\cref{fig:prompt_decompose})
    \item \textit{Visual evidence grounding}: fact statement extractor (\cref{fig:prompt_fact}), fact statement retrieval (\cref{fig:prompt_retrieval}), evidence navigation (\cref{fig:prompt_navigator})
    \item \textit{Visual-text statement verification}: statement verification via VLMs (\cref{fig:prompt_verification})
\end{itemize}

\noindent\textbf{Interval of Grounded moment} The grounded interval is determined by the anchor frame and direction navigation. For `\textit{look behind}', it starts at the anchor frame and ends at the video's end, while `\textit{look ahead}' starts at the video's beginning and ends at the anchor. For `\textit{look around}', a fixed 8-frame interval centered on the anchor frame is mapped back to the original video timestamp. Given the interval, we uniformly re-sample frames within the interval for VLM input, typically 8 or 16 frames, depending on the VLM's requirement.

\noindent \textbf{Computing resources.} Experiments are conducted on 4 NVIDIA-A6000 GPU and Azure Cloud APIs (for OpenAI models). The minimal GPU memory requirement is 24GB.
\\

\noindent \textbf{Reference}
\small
\begin{itemize}[leftmargin=2em]
    \item [{[1]}] Difei Gao, Ruiping Wang, Ziyi Bai, Xilin Chen. Env-QA: A Video Question Answering Benchmark for Comprehensive Understanding of Dynamic Environments. IEEE/CVF international conference on computer vision. 2021
\end{itemize}

\end{document}